\documentclass{article}

\usepackage[english]{babel}

\usepackage[letterpaper,top=2cm,bottom=2cm,left=3cm,right=3cm,marginparwidth=1.75cm]{geometry}

\usepackage{amsmath}
\usepackage{authblk}
\usepackage{graphicx}
\usepackage{array}
\usepackage{natbib}
\usepackage{multirow}
\usepackage{soul} 
\usepackage{xcolor}
\newcolumntype{C}[1]{>{\centering\arraybackslash}p{#1}}
\usepackage[colorlinks=true, allcolors=blue]{hyperref}

\definecolor{seapurple}{RGB}{112,48,160}

\title{Understanding Contextual Bias in Agricultural CNNs Through Hidden-Neuron Concept Induction}
\author[2]{Abhilekha Dalal\thanks{Corresponding author} (\texttt{abhilekha.dalal@wright.edu})}

\author[1]{Avishek Das
(\texttt{avishek@ksu.edu})}

\author[1]{Samatha Ereshi Akkamahadevi 
(\texttt{samatha94@ksu.edu})}

\affil[1]{Computer Science, Kansas State University, Manhattan, KS, USA}
\affil[2]{Computer Science and Engineering, Wright State University, Dayton, OH, USA}

\begin{document}
\maketitle

\begin{abstract}
Convolutional Neural Networks (CNNs) can classify images using background-dominated patches that contain little or none of the target object, showing that they exploit contextual information. Previous studies detected this behavior through classification accuracy but say little about what CNN hidden layers represent. We investigate whether Concept Induction can reveal interpretable visual properties associated with hidden-neuron activations on contextual image patches. Using five agricultural datasets with high, moderate, and low contextual bias, we analyze four CNN architectures at three network depths and five patch locations. Quantitative image descriptors of each patch are encoded in a knowledge graph and used as background knowledge for Concept Induction, which derives logical expressions characterizing neuron activation. These expressions are tested on held-out images using the Mann-Whitney U test and Cliff’s delta. Of 48,013 tested neuron–feature pairs, 13,164 (27.4\%) were validated. Validation was most frequent at intermediate depths (mean 40.2\% at mid layers versus 12.0\% at deep layers), and no patch location dominated across datasets. Among the tested descriptors, edge, Gabor, and Tamura texture features had the highest validation rates. Stronger contextual bias did not correspond to higher neuron–feature validation rates: moderately biased Sugarcane had the highest rate (40.5\%) and highly biased Cassava the lowest (19.2\%). These results suggest that behavioral context dependence and the semantic interpretability of hidden neurons capture different aspects of CNN behavior, and that Concept Induction can complement patch-based bias detection.

\end{abstract}

\section{Introduction}
\label{intro}

Convolutional Neural Networks (CNNs) have achieved remarkable performance across a wide range of image-classification tasks, establishing them as a fundamental component of modern computer vision systems~\citep{krizhevsky2012imagenet, he2016deep}. However, strong predictive performance does not necessarily imply that a model has learned the visual characteristics intended by the classification task. Deep neural networks can exploit unintended correlations and other predictive shortcuts present in the training data, achieving high performance without relying primarily on the characteristics of the target object~\citep{geirhos2020shortcut,ilyas2019adversarial, shahroudnejad2021survey}. Because CNNs learn statistical patterns directly from training data, such shortcuts can arise from contextual information, including background patterns, object co-occurrences, acquisition artifacts, and other incidental visual cues that correlate with class labels rather than from the object's genuine defining features~\citep{beery2018recognition,xiao2020noise}. Evidence of such behavior has been reported across diverse application settings, demonstrating that reliance on contextual or background information is not restricted to a particular dataset or domain~\citep{beery2018recognition, ross2024shortcut}. Such reliance can become problematic when these associations are specific to the training distribution, as changes in background composition, acquisition conditions, or other contextual characteristics may affect model robustness and generalization~\citep{geirhos2020shortcut, aniraj2023masking, ross2024shortcut}.

Standard evaluation measures, such as accuracy, precision, recall, and F1-score, quantify how successfully a model predicts class labels, but provide limited information about the visual evidence underlying those predictions. Consequently, a model that relies on unintended contextual cues may still achieve strong performance when the same correlations are present in the training and evaluation data\citep{shetty2019not, ribeiro2016should, zech2018variable}. This limitation has motivated diagnostic approaches that systematically modify the visual information available to a trained model, including object masking, background manipulation, image occlusion, cropping, and region- or patch-based evaluation\citep{aniraj2023masking, beery2018recognition, okonoda2026unmasking, martinez2026detection}. By restricting or altering portions of an image and observing the resulting model behavior, such approaches can reveal whether predictions remain supported when expected object information is reduced or contextual information is modified. In particular, spatial patch-based evaluation provides a direct means of examining whether restricted image regions retain sufficient information for classification. Predictive performance substantially above the expected chance level on such patches can therefore provide evidence that a model has learned class-discriminative information from spatially localized image content, including contextual regions that may not contain the complete target object\citep{ shamir2008evaluation, martinez2026detection,dalal2026evaluating}.

Building on this form of diagnostic evaluation, our prior studies systematically investigated contextual dependence by evaluating CNNs on spatially restricted image regions across different classification settings. Experiments involving biomedical images, a manually curated object-classification dataset, and agricultural image datasets showed that trained CNNs can retain non-trivial classification capability even when predictions are made from only a portion of the original image~\citep{okonoda2026unmasking, martinez2026detection, dalal2026evaluating}. In the agricultural setting, five predefined spatial regions such as top-left, top-right, center, bottom-left, and bottom-right were evaluated across multiple datasets and CNN architectures, revealing substantial differences in the extent to which models could classify images from restricted regions~\citep{dalal2026evaluating}. These findings provided a behavioral characterization of context dependence and demonstrated that its severity varies across datasets and model architectures. 


Patch-based evaluation, however, describes \emph{whether} predictive information is present within restricted image regions without explaining \emph{what} visual information the network internally represents from those regions. Classification performance alone does not reveal which hidden neurons respond to properties of the patches, how such representations vary across network depth, or which visual characteristics are associated with their activation. Consequently, identifying context-dependent classification behavior represents only one part of understanding how contextual information is encoded within a trained CNN. Examining the internal representations associated with these image regions is necessary to move from behavioral observations of context dependence toward a semantic characterization of the learned representations.

Existing interpretability methods provide several ways to examine such internal representations. Attribution and localization methods identify image regions contributing to predictions~\citep{gradcam_batra, simonyan2014deep, zhou2016learning}, while concept- and neuron-level approaches such as TCAV~\citep{kim2018tcav}, ACE~\citep{ghorbani2019ace}, Network Dissection~\citep{bau2017network}, and CLIP-Dissect~\citep{oikarinen2023clip} associate learned representations with human-interpretable concepts. However, these approaches generally characterize internal representations through attribution, concept association, or semantic labeling rather than logically structured, compositional descriptions of neuron activation grounded in quantitative visual properties or support reasoning over such properties using explicit background knowledge. This motivates a complementary approach in which hidden-neuron activation patterns are connected to structured symbolic representations, and formal concept learning characterizes the visual conditions associated with neuron responses~\cite{dalal2024understanding}.

In this work, we investigate the hidden-neuron representations associated with the contextual image patches used in our preceding agricultural context-bias analysis. We develop a neuro-symbolic analysis framework that integrates hidden-layer activation analysis with quantitative image descriptors and Description Logic Concept Induction. For selected hidden layers of four CNN architectures, neuron activations are extracted from predefined spatial image patches, while complementary visual characteristics of the same patches are represented using UDAT image descriptors. These descriptors are discretized and encoded within OWL knowledge graphs, providing structured background knowledge for Concept Induction. DL-Learner is then used to derive Description Logic concepts characterizing the visual properties associated with selected neuron activation patterns. The induced associations are subsequently evaluated on held-out images using statistical significance and effect-size criteria, providing a systematic mechanism for identifying semantically validated neuron--feature associations.

Using this framework, we examine representations across five agricultural image datasets selected to span the high-, moderate-, and low-context-bias conditions identified in the preceding patch-based study. The analysis considers multiple CNN architectures, three representative network depths, five spatial patch locations, and complementary families of quantitative image descriptors. This experimental design allows us to investigate not only whether semantic associations can be identified for hidden neurons, but also where these associations emerge within CNN feature hierarchies, how they vary spatially, which visual properties most frequently characterize them, and whether their prevalence corresponds to the degree of context dependence observed behaviorally.


Specifically, we investigate the following research questions:

\begin{itemize}
    \item \textbf{RQ1:} At which depths of CNN architectures do semantically interpretable neuron representations associated with contextual image patches emerge? At which layers of CNN architectures do semantically interpretable, context-sensitive representations emerge?
    
    \item \textbf{RQ1.1:} Which spatial patch locations consistently produce semantically validated neuron representations across agricultural datasets?
    
    \item \textbf{RQ2:} Which image-feature families most consistently characterize the hidden-neuron representations associated with contextual image patches?
    
    \item \textbf{RQ3:} How does the emergence of validated semantic concepts relate to the degree of context bias observed across agricultural datasets?
\end{itemize}

The main contributions of this work are as follows:

\begin{itemize}

    \item We extend the analysis of context dependence in agricultural image classification from observable model behavior to internal CNN representations through a neuro-symbolic framework that associates hidden-neuron activation patterns with semantically interpretable visual properties using Description Logic Concept Induction.

    \item We introduce a systematic analysis and validation pipeline for identifying neuron--feature associations, combining targeted neuron selection, symbolic concept generation, and independent statistical validation on held-out images.

    \item We provide a multi-dimensional empirical characterization of semantically validated representations across five agricultural datasets, four CNN architectures, three representative network depths, five spatial patch locations, and multiple visual feature families, revealing how these associations vary across depth, spatial region, and visual characteristics.

    \item We relate representation-level semantic validation to previously observed behavioral context dependence and show that stronger patch-based context bias does not necessarily correspond to a greater prevalence of semantically validated neuron--feature associations, highlighting that behavioral context dependence and semantic concept emergence capture distinct aspects of CNN behavior.

\end{itemize}

The remainder of this paper is organized as follows. Section~\ref{literature} reviews related work on contextual bias, CNN interpretability, and concept-based and neuro-symbolic explanation methods. Section~\ref{prelim} introduces the background concepts and technical foundations necessary to understand the proposed framework. Section~\ref{method} describes the datasets, hidden-layer analysis, symbolic representation construction, Concept Induction, and statistical validation framework. Section~\ref{result} presents and discusses the experimental findings with respect to the research questions. Finally, Section~\ref{conclude} summarizes the main findings, discusses limitations and implications, and outlines directions for future work.

\section{Related Work}
\label{literature}

\subsection{Context bias and shortcut learning in image classification}




Image classifiers can learn regularities of a dataset instead of properties of the target object, a problem  recognized even before deep learning became dominant. Early investigation showed that popular recognition benchmarks have their own distinctive signature~\citep{torralba2011unbiased}. A classifier could tell which dataset an image came from, and models trained on one dataset generalized poorly to others; such failures were later described as \ emph {shortcut learning}~\citep{geirhos2020shortcut}. A shortcut is a decision rule that works on the benchmark but fails when conditions change because it relies on spurious cues that only happen to correlate with the label.

Background reliance is the best documented shortcut in natural images. Camera-trap classifiers lose much of their accuracy at locations not seen during training~\citep{beery2018recognition}. Images with the object removed still yield non-trivial accuracy on ImageNet~\citep{xiao2020noise}. One study found a model that recognized aeroplanes from artificial padding patterns and horses from a source watermark~\citep{lapuschkin2019unmasking}. Similar reliance on unintended cues has been observed in medical imaging, where CNNs could identify the hospital 
that produced a chest radiograph with almost perfect accuracy~\citep{zech2018variable}. Since pneumonia rates differed between hospitals, this site information leaked into the pneumonia predictions. Likewise, COVID-19 detectors were also found to rely on confounders rather than lung pathology, leading to poor generalization across hospitals~\citep{degrave2021ai}.
In agricultural imaging, a model trained on only eight background pixels of PlantVillage images reached 49\% accuracy, compared with a chance level of 2.6\%~\citep{noyan2022uncovering}. Similarly, plant phenotyping models were found to make confounder-driven decisions, and correcting them required expert feedback during training~\citep{schramowski2020making}.

A related line of work measures dataset bias by classifying image regions that contain little or no task-relevant information. Small, seemingly blank sub-images extracted from object recognition benchmarks were classified with accuracy well above chance~\citep{model2015comparison}. Similar findings were reported in biomedical imaging, where CNNs achieved high classification accuracy using image regions without biomedical content~\citep{dhar2021evaluation}.

Building on this direction, subsequent studies investigated contextual bias across different image-classification domains. Image transformations were used to distinguish background bias from useful contextual information across several image-classification benchmarks~\citep{erukude2024identifying}. An investigation tested thirteen cancer pathology datasets with four CNN architectures~\citep{okonoda2026unmasking}. Background regions without clinical content were classified with up to 93\% accuracy. More recently, contextual bias was examined across eight agricultural image datasets, where background-dominated patches achieved above-chance classification accuracy in six datasets~\citep{dalal2026evaluating}. Although class imbalance explained part of this effect, it did not fully account for the results.

While these studies show that context bias appears in several domains through classification performance, , they do not investigate how the visual characteristics of contextual regions are reflected in CNN hidden-neuron activations. The present work addresses this gap by applying Concept Induction to analyze the relationships between visual properties of image patches and hidden-neuron activations, using agricultural datasets exhibiting high, moderate, and low levels of contextual bias identified in~\citep{dalal2026evaluating}.

\subsection{Explainability of CNN hidden representations}
Early work on CNN interpretability focused on visualizing the patterns that activate individual units. Deconvolutional networks were used to project intermediate activations back into pixel space~\citep{zeiler2014visualizing}, while gradient-based approaches generated class-maximizing images and saliency maps~\citep{simonyan2014deep}. Subsequent work introduced regularization techniques to make such visualizations easier to recognize~\citep{yosinski2015understanding}. These studies provided insights into the hierarchical representations learned by CNNs, where early layers respond to low-level features such as edges and colors, while deeper layers respond to more complex patterns such as object parts and objects. However, interpreting these visualizations remains largely qualitative.  Attribution methods such as Grad-CAM highlight the image regions contributing to a particular prediction~\citep{gradcam_batra}. They show \emph{where} the model looks, but not \emph{what} it has learned to represent.

Network Dissection introduced a quantitative approach to interpreting individual neurons by comparing thresholded activation maps with pixel-level annotations of colors, textures, materials, parts, objects, and scenes~\citep{bau2017network}. This approach was subsequently extended to logical combinations of concepts, revealing both abstract neurons and neurons that respond to several unrelated concepts~\citep{mu2020compositional}.

Other methods investigate concepts represented in the activation space rather than focusing exclusively on individual neurons. TCAV learns a concept direction from example images and measures its effect on a class prediction~\citep{kim2018tcav}, while ACE identifies concepts automatically by clustering image segments in activation space~\citep{ghorbani2019ace}. More recent methods use vision-language models to generate natural-language descriptions of hidden neurons~\citep{oikarinen2023clip,hernandez2022milan}. Neuron-level analysis has also been extended to shortcut learning, where individual neurons can be scored by how strongly they depend on spurious features~\citep{le2025xai}.

Despite these advances, existing approaches have limitations in the context of our study.  Concept-based explanations can depend on the probe dataset and the available concept vocabulary, potentially producing different interpretations under different settings~\citep{ramaswamy2023overlooked}. Traditional Dissection-style methods need pixel-level concept masks, which are generally unavailable in agricultural image datasets. Similarly, labels produced by vision-language models come from an opaque model themselves, so they inherit its lack of transparency. Finally, all of these methods provide labels, vectors, or scores; they do not generally induce logical statements grounded in explicit background knowledge. Logical statements of this kind would express the relationships between visual features and neuron activations to be represented as formally defined concepts that can be investigated within a symbolic reasoning framework.

\subsection{Neuro-symbolic explanation and Concept Induction}
Neuro-symbolic AI combines neural learning with symbolic knowledge representation and reasoning~\citep{hitzler2020neural, lee2024neural}. In explainable AI, ontologies can provide a structured vocabulary for interpreting neural network behavior. For example, classifiers have been trained to map a neural network's activations to ontology concepts, allowing model outputs to be explained using symbolic terms~\citep{desousaribeiro2021aligning}.

Concept Induction provides an alternative approach in which explanations are learned directly as logical class expressions. Given a knowledge base expressed in Description Logic (DL), along with sets of positive and negative individuals, the objective is to identify a class expression that covers the positive examples while excluding the negative examples~\citep{DBLP:journals/ml/LehmannH10,sarker2019efficient}. The resulting DL concepts have formal semantics and can express complex relationships through combinations of classes and properties using logical constructors such as conjunction, disjunction, and quantifiers. These expressions are interpreted with respect to the underlying background knowledge.

DL-Learner is one of the most established systems for concept induction that supports OWL and RDF background knowledge and provides access to reasoning-based learning algorithms, including refinement-based learners such as CELOE~\citep{lehmann2009dllearner,buhmann2016dllearner}. However, refinement-based approaches can be computationally expensive due to the large number of reasoning operations required during concept search. To address this limitation, Efficient Concept Induction from Individuals (ECII) was introduced as a computationally efficient alternative, although with a more restricted search space~\citep{sarker2019efficient}.

Concept Induction has previously been explored for explaining neural network predictions and internal representations. A Semantic Web-based approach demonstrated how induced concepts could be used to explain the classifications of trained neural networks~\citep{sarker2017explaining}, while a Wikipedia-derived class hierarchy was used to generate meaningful explanations of differences between data sets in ways that people found meaningful~\citep{widmer2023towards}.  This direction was subsequently extended to hidden-neuron analysis, where images that strongly activate a neuron were treated as positive examples and those producing weak activations as negative examples~\citep{dalal2024value,dalal2026toward}. The induced concepts are then validated against new, purpose-retrieved images to confirm that they describe the neuron’s behavior. This pipeline produced meaningful labels for dense-layer neurons of a scene classifier. The framework was subsequently automated~\citep{akkamahadevi2024automating} and later studies utilized this with different training targets and on the SUN2012 dataset~\citep{akkamahadevi2025evaluation,sensarma2026case}.


Previous applications of Concept Induction to hidden-neuron analysis have two limitations. Its background knowledge came from object annotations of whole images, and it analyzed only the last dense layer. The resulting concepts therefore say \emph{which objects} a neuron activation responds to, but not \emph{which visual properties} drive it, or how these change with network depth and image position. We address both of these limitations by building background knowledge from quantitative descriptors of contextual patches, and applying Concept Induction at three network depths.

\subsection{Image descriptors and semantic background knowledge}
Before the widespread adoption of deep learning, images were usually described with handcrafted descriptors that have exact mathematical definitions. Haralick features capture texture statistics derived from gray-level co-occurrence matrices (GLCMs), including measures such as contrast, correlation, and entropy~\cite{haralick1973textural}, while Tamura features approximate perceived texture properties, such as coarseness, contrast, and directionality~\cite{tamura1978textural}. Gabor filters measure texture patterns across several scales and orientations~\cite{manjunath1996texture}, whereas Local binary patterns describe fine texture and are invariant to gray-scale changes and rotation~\cite{ojala2002multiresolution}. These descriptors have explicit mathematical definitions and correspond to visual properties that can be named. A single value, such as high entropy, can therefore be read directly, unlike a learned CNN feature.

WND-CHARM extended the use of handcrafted features by introducing a comprehensive set of image descriptors applicable across different image domains~\cite{orlov2008wndcharm,shamir2008wndchrm}.
It extracts features from the raw image and its Fourier, wavelet, and Chebyshev transforms, including combinations of these transformations. The descriptor families include polynomial decompositions (Zernike and Chebyshev), texture features (Haralick, Tamura, and Gabor), edge and object statistics, and multiscale histograms. Together, these descriptors characterize different aspects of image appearance, including texture, frequency content, structure, and intensity distribution. The set was designed to work across image domains without tuning. UDAT builds on these descriptors as a general quantitative analysis tool~\cite{shamir2017udat}. Previous research has demonstrated that handcrafted and learned features can provide complementary information~\citep{nanni2017handcrafted}, while ImageNet-trained CNNs have been shown to exhibit a strong reliance on texture information~\citep{geirhos2018imagenettrained}. These findings motivate the use of quantitative image descriptors to investigate the visual characteristics associated with CNN hidden-neuron activations.

While handcrafted descriptors have traditionally been used for image classification and quantitative analysis, we employ them to construct symbolic background knowledge for explaining CNN hidden-neuron activations. In this work, UDAT converts low-level visual characteristics of each patch into structured descriptors, which are subsequently discretized and represented in OWL. This representation enables Concept Induction to identify logical descriptions of the visual properties associated with neuron activation patterns.

\section{Preliminaries}
\label{prelim}
\subsection{Knowledge graphs and Description Logic representation}

Knowledge graphs provide a structured representation of entities and their relationships, typically expressed as nodes connected through labeled relations. When supported by an ontology, a knowledge graph additionally provides a formal vocabulary for describing the types of entities, their properties, and the relationships that may hold among them. This representation enables data from individual instances to be combined with formally defined domain knowledge and supports logical reasoning over the resulting structure~\citep{DBLP:books/crc/Hitzler2010,DBLP:journals/cacm/Hitzler21}.

An ontology-based knowledge graph can be viewed as consisting of a terminological component (TBox) and an assertional component (ABox). The TBox defines the classes and properties used to describe a domain and the relationships among these classes, while the ABox contains assertions about individual instances. For example, an image may be represented as an individual of the class \texttt{Image} and connected through an object property to another individual representing one of its visual characteristics.

Description Logics (DLs) provide a formal knowledge-representation framework underlying ontology languages such as OWL. In addition to atomic classes and properties, DLs allow complex class expressions to be constructed using logical operators such as conjunction ($\sqcap$), disjunction ($\sqcup$), negation ($\neg$), and existential ($\exists$) and universal ($\forall$) restrictions~\citep{DBLP:books/crc/Hitzler2010}. For example, an expression of the form
\[
    \texttt{Image} \sqcap \exists\,\texttt{hasFeature}.\texttt{FeatureRange}
\]
describes images that are related through \texttt{hasFeature} to an instance belonging to \texttt{FeatureRange}. Such expressions provide a symbolic representation that can be automatically reasoned over and form the basis for the Concept Induction approach used in this work.

\subsection{Concept Induction}

Concept Induction~\citep{DBLP:journals/ml/LehmannH10} is a symbolic learning approach that derives Description Logic class expressions from positive and negative examples represented within a knowledge base. Given a set of positive examples $P$, a set of negative examples $N$, and a knowledge base $K$, Concept Induction searches for a class expression $E$ that characterizes the positive examples while distinguishing them from the negative examples. Ideally, an induced expression satisfies
$K\models E(p)$ for all $p \in P$
while
$K \not\models E(q)$ for all $q \in N$.
When perfect separation is not possible, candidate expressions can instead be evaluated according to how accurately they cover the positive and negative examples.

In this work, Concept Induction is performed using DL-Learner~\citep{buhmann2016dllearner}, a framework for learning Description Logic concepts from examples and structured background knowledge. Specifically, we employ its CELOE (Class Expression Learning for Ontology Engineering)~\citep{lehmann2011celoe} algorithm, which searches the space of possible class expressions using refinement operators and evaluates candidate expressions according to their ability to distinguish positive from negative examples. The resulting expressions can combine classes and relationships defined in the underlying ontology, producing symbolic descriptions of characteristics shared by the positive examples.

Within our analysis framework, this capability provides the connection between numerical CNN activation behavior and symbolic descriptions of image characteristics. Images exhibiting contrasting neuron activation behavior can be represented as positive and negative examples, while their visual characteristics are encoded in an OWL knowledge graph. Concept Induction can then identify class expressions describing visual properties associated with the neuron's activation behavior. The construction of these examples and the application and validation of the induced concepts are described in Section~\ref{method}.

\subsection{UDAT image descriptors}

UDAT~\citep{shamir2017udat} is a quantitative image analysis tool based on WND-CHARM~\citep{orlov2008wndcharm,shamir2008wndchrm} that extracts a broad range of numerical descriptors characterizing image content. These descriptors are computed from the original image as well as from image transformations, including Fourier, wavelet, Chebyshev, edge, color, and hue transforms, and combinations of these transformations(e.g., the Fourier transform of the wavelet transform). For each contextual image patch, UDAT generates 4,041 numerical descriptors covering different aspects of image appearance, including texture, intensity distribution, edges, and structural patterns.

In this study, we use only descriptors computed on the raw image (338 descriptors per patch, organized into 13 feature families). Descriptors of transformed images describe the image in a different domain (e.g., frequency coefficients of an edge map) and therefore cannot be related directly to visual properties of the patch. Whereas raw-image descriptors measure properties such as texture, edges, or intensity distribution that remain interpretable in the resulting explanations. Table~\ref{tab:udat-families} summarizes the feature families, the number of numerical descriptors associated with each family, and the visual properties they quantify. Together, they cover intensity and color distributions, texture, edges and objects, and frequency- and shape-related content. Each numerical descriptor is treated as an individual visual attribute for subsequent discretization and knowledge graph construction. The selection of feature families for individual CNN layers and their use in Concept Induction are described in Section~\ref{method}.

\begin{table}[t]
\centering
\caption{UDAT feature families considered in this study, the number of numerical descriptors associated with each family, and the visual properties they quantify.}
\label{tab:udat-families}
\small
\begin{tabular}{|p{3.4cm}|c|p{8.6cm}|}
\hline
\textbf{Feature family} & \textbf{\#Values} & \textbf{Visual property quantified} \\ \hline
Intensity statistics & 8 & Global distribution of pixel intensities (e.g., mean, median, standard deviation, entropy) \\ \hline
Color histogram & 19 & Distribution of colors in the image \\ \hline
Combined first four moments & 48 & Mean, standard deviation, skewness and kurtosis of intensity over image regions \\ \hline
Multiscale histograms & 24 & Intensity histograms computed at several bin resolutions \\ \hline
Edge statistics & 28 & Magnitude and direction of intensity gradients (means, medians, variances, histograms, homogeneity) \\ \hline
Object statistics & 34 & Properties of segmented objects: count, area, distances, Euler number, centroid \\ \hline
Haralick texture & 28 & Statistics of the gray-level co-occurrence matrix, e.g., contrast, correlation, entropy~\cite{haralick1973textural} \\ \hline
Tamura texture & 6 & Perceptual texture properties: coarseness, contrast, directionality~\cite{tamura1978textural} \\ \hline
Gabor filters & 7 & Responses to oriented band-pass filters at several frequencies \\ \hline
Fractal features & 20 & Self-similarity and roughness of image structure \\ \hline
Radon transform & 12 & Intensity projections along multiple angles, capturing oriented structure \\ \hline
Chebyshev--Fourier & 32 & Coefficients of orthogonal polynomial expansions describing spatial-frequency content \\ \hline
Zernike moments & 72 & Orthogonal moments describing shape and spatial distribution of intensity \\ \hline
\textbf{Total} & \textbf{338} & \\ \hline
\end{tabular}
\end{table}

\subsection{Prior patch-based context-bias study}
\label{sec:prior}

This study builds directly on a previous empirical investigation of context bias in CNN-based agricultural image classification~\citep{dalal2026evaluating}. The earlier work evaluated whether CNN classifiers could retain predictive performance when semantically meaningful content was largely removed from the input images. The study covered eight widely used plant and agricultural image datasets: PlantVillage~\citep{plantvillage}, PlantDoc~\citep{plantdoc}, DeepWeeds~\citep{DeepWeeds2019}, Cassava Leaf Disease~\citep{cassava}, WeedCrop~\citep{weedCrop}, OpenSprayer~\citep{openSprayer}, Sugarcane Leaf Disease~\citep{sugarcane}, and Oxford-102 Flowers~\citep{oxford102}. These datasets span plant disease recognition, weed identification, crop classification, spraying-status recognition, and fine-grained flower classification, and include both controlled and field-acquired imagery.
Four CNN architectures were evaluated: ResNet50V2, InceptionV3, DenseNet121, and VGG16. For each dataset, the models were trained on the original images and subsequently evaluated on predefined spatial image patches. Five spatial regions were considered for every image: top-left, top-right, center, bottom-left, and bottom-right. Each patch had an original size of (20 $\times$ 20) pixels and was resized to the input dimensions required by the corresponding CNN architecture. The original class label was preserved for each extracted patch, enabling direct evaluation of whether the trained classifier could predict the original class using only a spatially restricted region of the image. Most patches contained little or no agriculturally relevant visual content, although the center patch could occasionally include part of the primary subject.

Context dependence was assessed by comparing classification accuracy on the patch datasets with the corresponding random-chance accuracy of each classification task. For dataset $d$ containing $K_d$ classes, the uniform random-chance accuracy is:
\begin{equation}
C_d = \frac{1}{K_d}.
\end{equation} 
Patch performance substantially above this baseline was interpreted as evidence that the CNN could exploit contextual or dataset-specific signals even when the primary agricultural object or disease evidence was absent. Such signals may include background texture, lighting patterns, framing characteristics, acquisition artifacts, or other recurring visual regularities associated with the class labels.

\begin{table}[!ht]
\centering
\caption{Summary of dataset characteristics and patch-based context-bias results reported in the previous study. ``Chance Acc." denotes random-chance accuracy, and ``Patch Acc." denotes the reported patch accuracy range.}
\label{tab:prev_dataset_comparison}
\resizebox{\textwidth}{!}{%
\begin{tabular}{|
C{1.8cm}|
C{2.0cm}|
C{1.3cm}|
C{2cm}|
C{1.2cm}|
C{1.2cm}|
C{1.6cm}|
C{1.2cm}|
C{1.7cm}|}
\hline
\textbf{Dataset} &
\textbf{Tasks} &
\textbf{\# Images} &
\textbf{Acquisition Setting} &
\textbf{\# Classes} &
\textbf{Random Chance} &
\textbf{Patch Acc. Range} &
\textbf{Margin} &
\textbf{Context-Bias Category} \\ \hline

PlantVillage &
Disease classification &
$\sim$54,000 &
Controlled &
15 &
6.7\% &
9.1–19.1\% &
6.2 pp & 
Moderate \\ \hline

PlantDoc &
Disease classification &
2,569 &
Field &
27 &
3.7\% &
2.0–7.1\% &
0.03 pp &
Low \\ \hline

DeepWeeds &
Species classification &
17,509 &
Field &
9 &
11.1\% &
51.3–55.0\% &
40.6 pp &
High \\ \hline

Cassava &
Disease classification &
21,397 &
Field  &
5 &
20\% &
16.0–55.8\% &
13.6 pp &
High \\ \hline

WeedCrop &
Crop/weed classification &
2,822 &
Field/UAV &
2 &
50\% &
85.6–92.4\% &
41.1 pp &
High \\ \hline

OpenSprayer &
Weed classification &
$\sim$10,000 &
UAV / aerial &
2 &
50\% &
58.4–80.6\% &
20.2 pp &
High \\ \hline

Sugarcane &
Disease detection &
$\sim$3,000+ &
Field  &
11 &
9.1\% &
11.4–22.8\% &
7.8 pp &
Moderate \\ \hline

Oxford 102 &
Species classification &
$\sim$8,200 &
Field &
102 &
0.98\% &
0.6–3.9\% &
0.7 pp &
Low \\ \hline
\end{tabular}
}
\end{table}

The previous analysis revealed considerable variation in context dependence across datasets. DeepWeeds, WeedCrop, OpenSprayer, and Cassava exhibited pronounced above-chance performance on patch-only inputs, whereas PlantVillage and Sugarcane showed more moderate context dependence. In contrast, PlantDoc and Oxford-102 Flowers produced patch accuracies close to their corresponding chance baselines, indicating comparatively lower dependence on restricted contextual regions. The study also found that susceptibility to context-dependent behavior varied across CNN architectures, demonstrating that architectural design alone does not guarantee robustness against spurious contextual correlations.
These findings established that patch-level predictive behavior can reveal the presence and relative degree of contextual dependence in CNN-based agricultural classifiers. However, the analysis remained primarily behavioral: it demonstrated that models could exploit restricted image regions, but did not identify which visual characteristics within those regions were represented by the CNNs or how such representations emerged across hidden layers. The previous study therefore identified feature-level investigation as an important direction for future work, including the use of image descriptors such as those provided by UDAT~\citep{shamir2017udat} to characterize the visual information contained in contextual patches.

The present study extends this earlier work from detecting context dependence to explaining its internal representation. Using the same patch-based experimental setting, we investigate hidden-neuron activations across multiple CNN depths and characterize them using UDAT-derived visual descriptors, knowledge graph representations, Description Logic Concept Induction, and held-out statistical validation. This enables analysis of not only whether contextual bias is present, but also which semantically interpretable visual feature families are associated with the corresponding hidden-neuron representations.




\section{Methodology}
\label{method}

\subsection{Neuro-Symbolic Analysis Framework}

\begin{figure}[!ht]
    \centering
    \includegraphics[width=\textwidth]{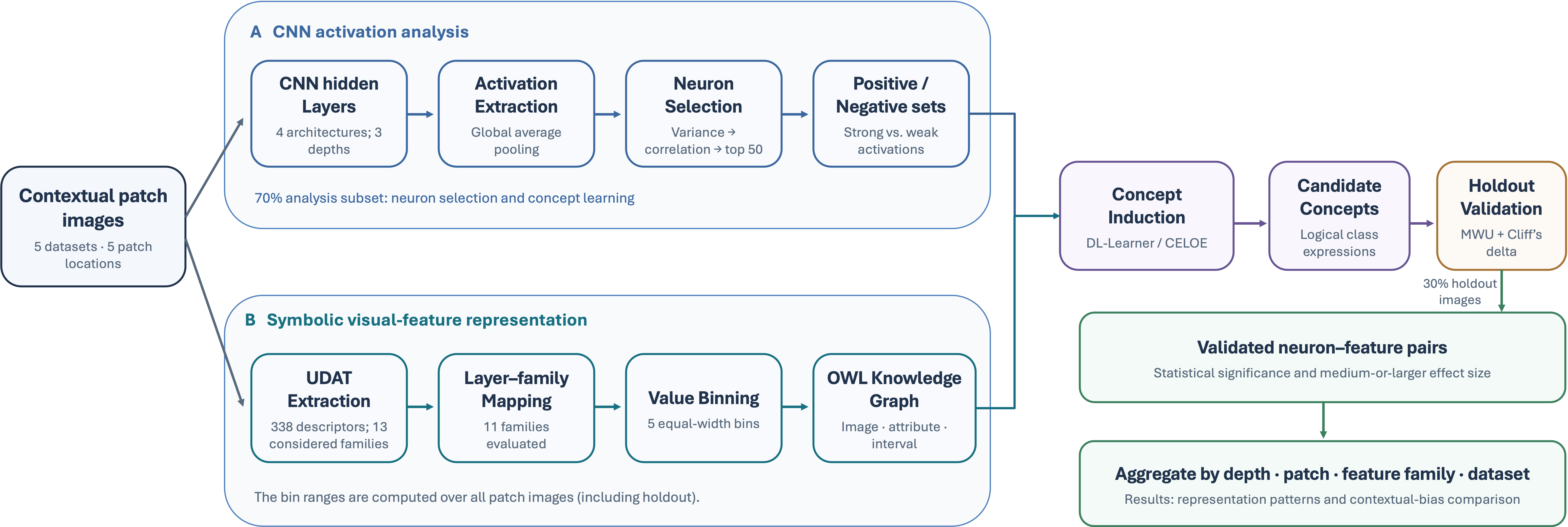}
    \caption{Overview of the neuro-symbolic analysis pipeline for interpreting CNN hidden-neuron activations from contextual image patches.}
    \label{fig:overall}
\end{figure}

This study adapts a neuro-symbolic hidden-neuron explanation approach based on Concept Induction~\citep{dalal2024value} 
to investigate the internal representations associated with contextual image patches. The analysis builds on the agricultural context-bias setting described in Section~\ref{sec:prior} and shifts the focus from observing patch-level classification behavior to characterizing the visual properties represented within CNN hidden layers.
Figure~\ref{fig:overall} presents an overview of the analysis pipeline. For each dataset and spatial patch location, patch images are provided to the corresponding trained CNN, and hidden-layer activations are collected at three network depths: mid, mid-deep, and deep. The resulting activation patterns are used to identify neurons suitable for subsequent explanation analysis.

In parallel, the visual characteristics of the patch images are extracted using UDAT image descriptors. The extracted feature values are transformed into symbolic representations and encoded in an OWL knowledge graph, which provides the background knowledge used for Concept Induction. For each selected neuron-feature-family combination, images associated with contrasting activation behavior are used as positive and negative examples. Concept Induction then searches for Description Logic expressions using the OWL knowledge graph that distinguish the positive examples from the negative examples, thereby generating candidate semantic explanations of neuron activation.
The resulting neuron-feature associations are evaluated using held-out data and statistical validation. Associations that satisfy the predefined validation criteria are retained as validated neuron-feature pairs. These pairs form the fundamental observations used in the subsequent analysis. Finally, the validated neuron-feature pairs are aggregated along several experimental dimensions, including CNN depth, spatial patch location, dataset, and image-feature family.

These aggregated representations are used to examine where context-sensitive semantic representations emerge within CNNs. In particular, the analysis examines the network depths at which such representations arise, the spatial regions and visual feature families with which they are most consistently associated, and how their occurrence relates to the context-bias levels observed in the preceding patch-based study.

The prior patch-based study categorized the eight datasets according to context-bias severity observed from patch-level classification performance. Since the subsequent hidden-neuron and Concept Induction analysis is considerably more computationally intensive, we selected a representative subset of five datasets rather than repeating the analysis across all eight datasets. The selection is designed to span the observed bias spectrum while providing greater representation of datasets exhibiting strong contextual dependence, for which identifying the internal representations associated with image context is particularly relevant. Accordingly, Cassava, WeedCrop, and OpenSprayer from the high-bias group, Sugarcane from the moderate-bias group, and Oxford-102 from the low-bias group have been chosen. This subset therefore enables comparison of hidden-layer semantic representations across datasets exhibiting substantially different levels of previously observed context dependence.

\subsection{Hidden-Layer Selection and Activation Extraction}

To examine how context-sensitive representations evolve across network depth, three hidden layers were selected from each CNN architecture to represent mid, mid-deep, and deep stages of feature processing. The layers are selected from progressively deeper major processing blocks within each architecture, providing representative points along the feature hierarchy while avoiding adjacent layers that capture highly similar stages of computation. Because ResNet50V2, DenseNet121, InceptionV3, and VGG16 differ substantially in their connectivity patterns and architectural organization, we do not assume exact layer-wise equivalence across the four networks. Instead, the selected layers provide architecture-specific reference points corresponding broadly to intermediate, later-intermediate, and deep feature-processing stages. The selected layers are summarized in Table~\ref{tab:cnn-layers}. 

\begin{table}[ht]
\centering
\caption{Hidden layers selected to represent mid, mid-deep, and deep stages of each CNN architecture}
\label{tab:cnn-layers}

\begin{tabular}{|C{3cm}|C{3.5cm}|C{3.5cm}|C{3cm}|}
\hline
\textbf{Architecture} & \textbf{Mid} & \textbf{Mid-deep} & \textbf{Deep} \\ \hline
ResNet50V2
& conv3\_block4\_out
& conv4\_block6\_out
& conv5\_block3\_out \\ \hline

DenseNet121
& conv3\_block12\_concat
& conv4\_block24\_concat
& relu \\ \hline 
InceptionV3
& mixed5
& mixed7
& mixed10 \\ \hline

VGG16
& block3\_conv3
& block4\_conv3
& block5\_conv3 \\ \hline
\end{tabular}
\end{table}

Earlier and intermediate convolutional layers primarily encode basic visual features, such as edges, colors, and textures, into local patterns, while progressively deeper layers combine information from larger image regions to represent more complex and abstract features~\citep{zeiler2014visualizing}.
Analyzing features at these different layer depths helps us understand how context-sensitive concepts gradually develop and change as information moves through the CNN.


For each dataset and spatial patch location, the corresponding patch images are forwarded through the trained CNN, and activation values are extracted from the selected hidden layers. Since the analyzed layers are convolutional or feature-map-producing layers, each layer contains a distinct set of feature channels. Consequently, neurons analyzed at the mid, mid-deep, and deep stages are not treated as the same neurons tracked across depth; rather, each selected layer contributes its own set of candidate neurons for subsequent analysis. For each selected layer, an intermediate model was constructed using the trained CNN's input and the selected hidden layer as its output. Since the analyzed layers produce multi-channel feature maps, global average pooling (GAP) was applied to reduce each channel's activation map to a single value by averaging over its spatial positions. Each feature channel is treated as a neuron in the subsequent analysis. Repeating this procedure over all patch images produced an image-by-neuron activation matrix for each combination of dataset, patch location, architecture, and selected hidden layer.

The extracted activation values are then used as the basis for neuron filtering and the construction of positive and negative example sets for Concept Induction. To avoid information leakage during concept evaluation, the constructed patch data are divided into analysis and held-out subsets, with the former used for neuron selection and concept generation and the latter for validation.

\subsection{Neuron Selection}
To restrict Concept Induction to neurons exhibiting informative and decision-relevant activation behavior, a three-stage neuron-selection procedure is applied independently to each analyzed hidden layer. Importantly, neuron selection is performed using only the 70\% analysis split; the 30\% holdout split was excluded from all filtering and concept-generation steps and reserved exclusively for subsequent validation. For each neuron, activation values are collected across the analysis images. The first filtering stage removed neurons exhibiting comparatively little variation in their responses. Neuron-wise activation variance was calculated, and neurons falling within the bottom 25th percentile of the variance distribution for the corresponding layer were discarded. Because this criterion is quantile-based rather than based on a fixed variance threshold, the cutoff adapts to the activation characteristics of each layer.
Among the remaining neurons, a second filtering stage assessed association with the CNN's prediction confidence. For each neuron \(n\), the Pearson correlation coefficient between its activation values and the corresponding model-confidence scores was calculated,

$$ r_n = \frac{\sum_i (a_{in}-\bar{a}_n)(c_i-\bar{c})} {\sqrt{\sum_i(a_{in}-\bar{a}_n)^2} \sqrt{\sum_i(c_i-\bar{c})^2}}, $$

where \(a_{in}\) denotes the activation of neuron \(n\) for image \(i\), and \(c_i\) denotes the CNN prediction confidence for that image. Neurons are ranked according to the absolute correlation magnitude, \(|r_n|\), and the 100 highest-ranked neurons are retained. Using the absolute value allows both positively and negatively associated activation patterns to be considered potentially relevant to model confidence. Finally, the retained 100 neurons are re-ranked according to their activation variance, and the 50 highest-variance neurons are selected for Concept Induction. Thus, each analyzed layer contributes at most 50 candidate neurons for each evaluated feature family.

\begin{table}[t]
\centering
\caption{Reduction in candidate neurons across the three-stage neuron-selection procedure for the analyzed CNN layers.}
\label{tab:neuron_selection_reduction}
\resizebox{\textwidth}{!}{%
\begin{tabular}{
    |C{2.2cm}|
    C{1.5cm}|
    C{3.5cm}|
    C{1.8cm}|
    C{1.8cm}|
    C{1.8cm}|
    C{1.4cm}|
}
\hline
\textbf{Architecture} &
\textbf{Depth} &
\textbf{Layer} &
\textbf{Initial neurons/ channels} &
\textbf{After variance filter} &
\textbf{After correlation ranking} &
\textbf{Final selected} \\
\hline

ResNet50V2 & Mid & \texttt{conv3\_block4\_out} & 512 & 384 & 100 & 50 \\ \hline
ResNet50V2 & Mid-deep & \texttt{conv4\_block6\_out} & 1,024 & 768 & 100 & 50 \\ \hline
ResNet50V2 & Deep & \texttt{conv5\_block3\_out} & 2,048 & 1,536 & 100 & 50 \\ \hline

DenseNet121 & Mid & \texttt{conv3\_block12\_concat} & 512 & 384 & 100 & 50 \\ \hline
DenseNet121 & Mid-deep & \texttt{conv4\_block24\_concat} & 1,024 & 768 & 100 & 50 \\ \hline
DenseNet121 & Deep & \texttt{relu} & 1,024 & 768 & 100 & 50 \\ \hline

InceptionV3 & Mid & \texttt{mixed5} & 768 & 576 & 100 & 50 \\ \hline
InceptionV3 & Mid-deep & \texttt{mixed7} & 768 & 576 & 100 & 50 \\ \hline
InceptionV3 & Deep & \texttt{mixed10} & 2,048 & 1,536 & 100 & 50 \\ \hline

VGG16 & Mid & \texttt{block3\_conv3} & 256 & 192 & 100 & 50 \\ \hline
VGG16 & Mid-deep & \texttt{block4\_conv3} & 512 & 384 & 100 & 50 \\ \hline
VGG16 & Deep & \texttt{block5\_conv3} & 512 & 384 & 100 & 50 \\ \hline

\end{tabular}
}
\end{table}

\subsection{UDAT Feature Extraction and Discretization}
\label{sec:udat}

Each contextual patch is described with UDAT, using the 338 raw-image descriptors organized into 13 feature families (Table~\ref{tab:udat-families}).

Not every family is tested against every hidden layer. Feature families are assigned to individual network layers based on an LLM-guided assessment. The assessment was conducted using 
Claude Sonnet (Anthropic, 2026). For each architecture, the model was prompted with the layer names, spatial resolutions, and channel depths alongside descriptions of the 13 UDAT feature families, and asked to classify each layer-feature-family combination as \emph{test}, \emph{optional/borderline}, or \emph{skip} based on the type of visual representations expected at that network depth. Architecture-specific properties were also considered, including residual skip connections in ResNet50V2, dense feature concatenation in DenseNet121, and multi-scale parallel processing in InceptionV3. The full prompt template is provided in Appendix~\ref{app:llm-assessment}. of their relevance to the type of visual representations expected at different network depths. of their relevance to the type of visual representations expected at different network
depths. The assessment categorized each layer-feature-family combination as test, optional/borderline, or skip, and only the \emph{test} combinations are run through Concept Induction. This gives 84 layer-feature-family combinations per dataset-patch configuration, the same for all 25 configurations. Of the 13 families, 11 are evaluated. Intensity Statistics and Color Histogram are marked \emph{test} only for early layers, which are not analyzed in this study. Zernike moments are run on the mid-deep and deep layers, but CELOE returns only the generic concept \texttt{Thing} for every neuron (7,759 of 7,759 runs that return a concept). These concepts fall under the exclusion rule described in the validation subsection and are not validated, so Zernike moments produce no testable concepts and do not appear in the results.

The raw UDAT outputs are continuous numerical values and therefore cannot be used directly as symbolic concepts in the subsequent Description Logic representation. Each descriptor is therefore discretized into a finite set of ordered value intervals, and each interval is represented as a symbolic bin. A patch is then described by the interval its measurement falls into, not by the exact value. Interval boundaries are set from the minimum and maximum of each descriptor over all patch images of the corresponding dataset-patch configuration, i.e., analysis and held-out images together.

For descriptor $f_j$ and patch image $x$, the discretization is

\begin{equation}
f_j(x) \mapsto B_{j,k},
\end{equation}

where $f_j(x)$ is the numerical value of the descriptor for $x$, and $B_{j,k}$ is the $k$-th interval of $f_j$, the one containing that value. For example, a low edge-statistics value, falls into a lower-order bin, and a comparatively high value for the same descriptor falls into a higher-order bin. The discretized representation provides two advantages for the neuro-symbolic analysis. First, it transforms continuous visual measurements into categorical concepts that can be represented explicitly in OWL. Second, it allows Concept Induction to search for human-interpretable combinations of visual characteristics rather than operating directly over raw numerical values.

\subsection{Knowledge Graph Construction}

The discretized UDAT descriptors are encoded as an OWL-based knowledge graph to provide the symbolic background knowledge required for Concept Induction. Each contextual patch image is represented as an individual of the class `Image'. For each UDAT descriptor, an object property connects the image individual to a bin individual representing its discretized feature value. The bin individual is subsequently typed using an OWL class corresponding to the numerical interval into which the original descriptor value fell.

\begin{figure}[t]
\centering
\includegraphics[width=\linewidth]{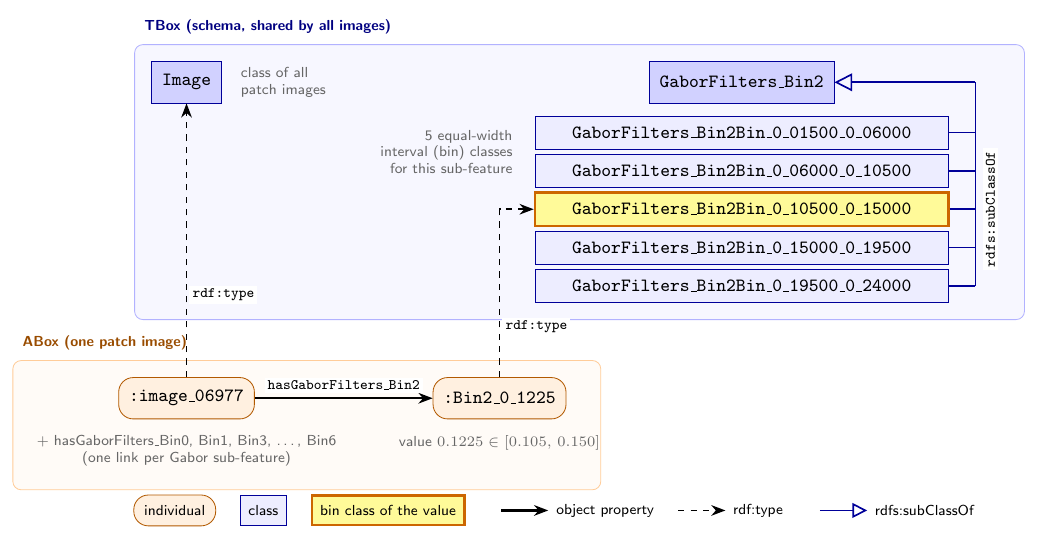}
\caption{Ontology pattern for one patch image and one UDAT sub-feature (Gabor filters,
sub-feature Bin2). The patch individual is linked by hasGaborFilters\_Bin2 to a value
individual (value 0.1225), which is typed with the interval class for $[0.105, 0.150]$;
the five equal-width interval classes are subclasses of the sub-feature class
GaborFilters\_Bin2. The same pattern is repeated for each of the seven Gabor sub-features.}
\label{fig:ontology}
\end{figure}

For example, in figure~\ref{fig:ontology}, for the Gabor feature family, properties such as hasGaborFilters\_Bin0, hasGaborFilters\_Bin1, ..., hasGaborFilters\_Bin6 connect an image to the discretized values of the corresponding Gabor descriptors. The interval-specific classes are organized under broader descriptor classes using rdfs:subClassOf, thereby preserving both the feature identity and its discretized numerical range. Formally, for image $x$ and descriptor $f_j$ with object property $p_j$ and interval classes $B_{j,1},\dots,B_{j,K}$, the graph contains the assertions
\begin{equation}
p_j(x, b), \qquad B_{j,k}(b),
\end{equation}
where $b$ is the value individual of $x$ for $f_j$, and $k$ is the index of the interval containing the numerical value $f_j(x)$, as defined in the discretization step.
Individual OWL representations generated for the patch images are subsequently combined into a single ontology for the corresponding dataset, patch location, and UDAT feature family. This combined knowledge graph is then supplied as the background knowledge in Concept Induction. The ontology therefore encodes each patch as a structured set of symbolic visual attributes. The discretized feature intervals are organized hierarchically, allowing Concept Induction to reason over both specific value ranges and broader descriptor-level abstractions.


\subsection{Concept Induction}

For each selected neuron--feature-family combination, Concept Induction is run with the CELOE algorithm in DL-Learner, using the OWL knowledge graph built from the corresponding UDAT descriptors as background knowledge. Only the 70\% analysis split is used; the 30\% holdout split remains unseen until the validation stage.
For each neuron $n$, positive and negative example sets are derived from its activations over the analysis images. Let $a_{in}$ denote the activation of neuron $n$ for image $x_i$, and let $a_n^{\max}$ be the maximum of $a_{in}$ over the analysis images. Images with activation of at least 80\% of $a_n^{\max}$ form the positive set, and images with nonzero activation of at most 20\% of $a_n^{\max}$ form the negative set:
\begin{align}
P_n &= \left\{\, x_i \mid a_{in} \geq 0.8\,a_n^{\max} \right\}, \\
N_n &= \left\{\, x_i \mid 0 < a_{in} \leq 0.2\,a_n^{\max} \right\}.
\end{align}
Images with zero activation, and images with activation between 20\% and 80\% of the maximum are discarded.
The two example sets and the OWL background knowledge are passed to DL-Learner as a \texttt{posNegStandard} learning problem with a closed-world reasoner. CELOE searches the space of Description Logic class expressions for those that cover the positive examples while excluding the negative ones. Each run is limited to 60~s.
For each neuron-feature-family combination, only the top-ranked class expression, ranked by predictive accuracy, is retained and carried forward to validation. A run for which CELOE returns no solution yields no candidate concept.

\subsection{Concept and Neuron-Feature Pair Validation}

The concepts generated from the 70\% analysis split are evaluated using the independent 30\% holdout split. Concepts containing only generic classes, such as \texttt{Thing} or \texttt{Image}, and concepts that cannot be parsed into UDAT feature-bin constraints are excluded from statistical validation.

For each processable neuron-feature pair, holdout images are divided into concept-positive and concept-negative groups based on whether they satisfy the induced concept. An image satisfies a concept when its UDAT feature values fall within the bin intervals specified by the corresponding Description Logic expression. For negated constraints, the feature values must fall outside the specified intervals. Images satisfying the concept are assigned to the concept-positive group, while the remaining images form the concept-negative group.

Statistical validation is performed only when both groups contain at least three images. A one-sided Mann-Whitney U test is applied to examine whether neuron activations are significantly higher for concept-positive images than for concept-negative images. The alternative hypothesis is defined as

\begin{equation}
H_1: A^{+}_n > A^{-}_n,
\end{equation}

where $A^{+}_n$ and $A^{-}_n$ denote the activation distributions of neuron $n$ for concept-positive and concept-negative holdout images, respectively. Statistical significance is assessed at $\alpha = 0.05$.

In addition to statistical significance, Cliff's delta ($\delta$) is used to measure the effect size of the activation difference between the two groups. Effect sizes are categorized as negligible ($|\delta| < 0.147$), small ($0.147 \leq |\delta| < 0.330$), medium ($0.330 \leq |\delta| < 0.474$), and large ($|\delta| \geq 0.474$). A neuron--feature pair is considered validated only if it satisfies both criteria: $p < 0.05$ and $|\delta| \geq 0.330$. This ensures that retained associations exhibit both statistically significant activation differences and at least medium effect-size magnitude.

The validation rate (VR) is calculated as the proportion of statistically tested neuron-feature pairs that satisfy the final validation criteria. For each CNN depth $d$ and dataset-patch configuration $c$, it is defined as

\begin{equation}
VR_{d,c} =
\frac{N^{\mathrm{validated}}_{d,c}}
{N^{\mathrm{tested}}_{d,c}}
\times 100,
\end{equation}

where $N^{\mathrm{validated}}_{d,c}$ denotes the number of validated neuron-feature pairs and $N^{\mathrm{tested}}_{d,c}$ denotes the number of pairs for which the Mann-Whitney U test is completed. Each configuration represents one dataset-patch combination, giving 25 configurations in total. The mean validation rate at depth $d$ is calculated as

\begin{equation}
\overline{VR}_{d} =
\frac{1}{25}
\sum_{c=1}^{25} VR_{d,c}.
\end{equation}

This aggregation gives equal weight to each dataset-patch configuration rather than pooling all neuron-feature pairs into a single rate.

For spatial analysis, validation results are aggregated across the analyzed architectures and layer depths for each dataset $s$ and patch position $p$. The corresponding patch-level validation rate is

\begin{equation}
VR_{s,p} =
\frac{N^{\mathrm{validated}}_{s,p}}
{N^{\mathrm{tested}}_{s,p}}
\times 100.
\end{equation}

The same validated-to-tested ratio is used for feature-family analysis, with neuron-feature pairs grouped according to their corresponding UDAT descriptor families. These measures provide the basis for the subsequent comparisons across CNN depths, spatial patch locations, and visual feature families.

\section{Results and Discussion}
\label{result}
\begin{figure}[!ht]
    \centering
    \includegraphics[width=\textwidth]{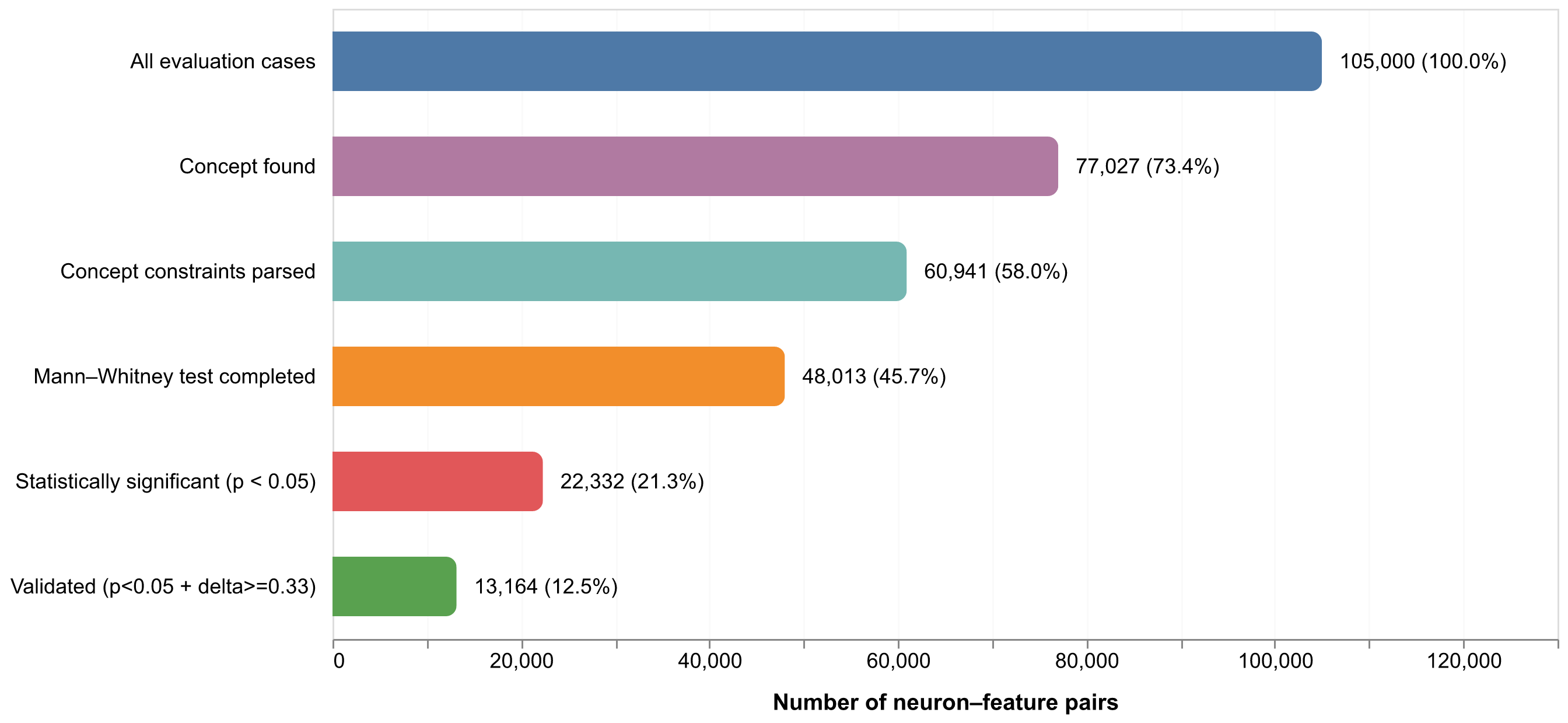}
    \caption{Progressive reduction of neuron-feature evaluation cases. Percentages indicate the proportion retained relative to the initial 105,000 evaluation cases.}
    \label{fig:prog_reduc}
\end{figure}

In this section, we first summarize the overall outcome of the neuron-feature validation pipeline. Across the 5 analyzed datasets and 5 spatial patch positions, the combined analysis comprised 25 dataset-patch configurations. For each configuration, 3 depths × 4 architectures = 12 hidden layers are examined, with 50 selected neurons evaluated for each applicable layer-feature-family combination. Because the availability of UDAT feature families varied across layers, 84 layer-feature-family combinations are represented per configuration, yielding \(84 \times 50 = 4{,}200\) neuron-feature evaluation cases. Across all 25 configurations, this resulted in a total of \(25 \times 4{,}200 = 105{,}000\) evaluation cases. Figure~\ref{fig:prog_reduc} summarizes the progressive reduction of neuron-feature evaluation cases through concept induction, constraint parsing, held-out statistical testing, and final validation. A total of 48,013 neuron-feature pairs satisfied the requirements for statistical testing on the held-out data. Among them, 13,164 pairs met the final validation criteria, corresponding to an overall validation rate of approximately 27.4\%. To examine whether this reduction is consistent across experimental settings, Table~\ref{tab:all_reduction} presents the validation progrssion separately for all 25 dataset-patch configurations. A progressive decrease can be observed in every configuration, indicating that the overall filtering and validation behavior is broadly distributed rather than dominated by a single dataset or spatial region.

The following sections analyze these validated neuron–feature pairs across CNN depth, spatial patch location, feature family, and dataset-level context-bias characteristics to address the research questions.

\begin{table}[!ht]
\caption{Dataset-patch-wise progression of neuron-feature evaluation cases. Patch locations are denoted as BL (Bottom-Left), BR (Bottom-Right), C (Center), TL (Top-Left), and TR (Top-Right).}
\label{tab:all_reduction}

\resizebox{\textwidth}{!}{%
\begin{tabular}{|l|c|c|c|c|c|c|c|}

\hline
\textbf{Dataset} &
  \textbf{Patch} &
  \textbf{All Cases} &
  \textbf{Concept Found} &
  \textbf{Constraints Parsed} &
  \textbf{MW Test Completed} &
  \textbf{Significant} &
  \textbf{Validated} \\ \hline
\multirow{5}{*}{Cassava}     & BL & 4200 & 3219 & 2356 & 2260 & 1200 & 362 \\ \cline{2-8} 
                             & BR & 4200 & 3184 & 2318 & 2223 & 1213 & 387 \\ \cline{2-8} 
                             & C  & 4200 & 3155 & 2241 & 2170 & 1131 & 433 \\ \cline{2-8} 
                             & TL & 4200 & 3227 & 2351 & 2296 & 1236 & 487 \\ \cline{2-8} 
                             & TR & 4200 & 3242 & 2350 & 2302 & 1266 & 495 \\ \hline
\multirow{5}{*}{OpenSprayer} & BL & 4200 & 3005 & 2367 & 2003 & 791  & 468 \\ \cline{2-8} 
                             & BR & 4200 & 3011 & 2282 & 1965 & 762  & 405 \\ \cline{2-8} 
                             & C  & 4200 & 3083 & 2422 & 2036 & 796  & 453 \\ \cline{2-8} 
                             & TL & 4200 & 3016 & 2286 & 1937 & 701  & 385 \\ \cline{2-8} 
                             & TR & 4200 & 3001 & 2367 & 2048 & 838  & 484 \\ \hline
\multirow{5}{*}{Oxford}      & BL & 4200 & 3081 & 2534 & 2248 & 1071 & 622 \\ \cline{2-8} 
                             & BR & 4200 & 3039 & 2463 & 2216 & 1043 & 667 \\ \cline{2-8} 
                             & C  & 4200 & 3243 & 2665 & 2259 & 824  & 423 \\ \cline{2-8} 
                             & TL & 4200 & 3059 & 2498 & 2211 & 1262 & 745 \\ \cline{2-8} 
                             & TR & 4200 & 3053 & 2449 & 2181 & 1074 & 650 \\ \hline
\multirow{5}{*}{Sugarcane}   & BL & 4200 & 2985 & 2248 & 2016 & 1132 & 895 \\ \cline{2-8} 
                             & BR & 4200 & 2946 & 2194 & 1961 & 1074 & 783 \\ \cline{2-8} 
                             & C  & 4200 & 3132 & 2438 & 2187 & 1045 & 650 \\ \cline{2-8} 
                             & TL & 4200 & 3016 & 2245 & 1916 & 1133 & 852 \\ \cline{2-8} 
                             & TR & 4200 & 2865 & 2178 & 1926 & 1098 & 877 \\ \hline
\multirow{5}{*}{WeedCrop}    & BL & 4200 & 3315 & 2973 & 1169 & 244  & 244 \\ \cline{2-8} 
                             & BR & 4200 & 3054 & 2707 & 993  & 184  & 184 \\ \cline{2-8} 
                             & C  & 4200 & 3020 & 2651 & 1303 & 562  & 562 \\ \cline{2-8} 
                             & TL & 4200 & 2960 & 2628 & 944  & 223  & 223 \\ \cline{2-8} 
                             & TR & 4200 & 3116 & 2730 & 1243 & 429  & 428 \\ \hline
\end{tabular}
}
\end{table}

\subsection{RQ1 - Emergence Across CNN Depth}

As the analysis is performed on the predefined patch regions used to assess contextual dependence, and the induced concepts are derived from visual properties of those same patches, the validated neuron-feature pairs are interpreted as context-sensitive internal representations. Table~\ref{tab:layer_depth_validation} summarizes the validation rates across different layer depths, including the mean, standard deviation, minimum, and maximum values.

\begin{table}[!ht]
\centering
\caption{Summary of neuron-feature validation rates across the three analyzed CNN depth levels.}
\label{tab:layer_depth_validation}
\resizebox{\textwidth}{!}{%
\begin{tabular}{
    |C{2.2cm}
    |C{2.5cm}
    |C{4.5cm}
    |C{1cm}
    |C{1.5cm}
    |C{1.8cm}|
}
\hline
\textbf{Layer depth} &
\textbf{Configurations} &
\textbf{Mean validation rate (\%)} &
\textbf{SD} &
\textbf{Min (\%)} &
\textbf{Max (\%)} \\
\hline
Mid      & 25 & 40.21 & 10.38 & 24.49 & 61.83 \\ \hline
Mid-deep & 25 & 26.91 & 12.89 & 14.43 & 54.02 \\ \hline
Deep     & 25 & 11.99 & 5.65  & 4.41  & 29.07 \\
\hline
\end{tabular}
}
\end{table}

Across the analyzed configurations, the Mid layers exhibit the highest mean validation rate at 40.21\%, followed by the Mid-deep layers at 26.91\%, while the Deep layers exhibit a substantially lower mean rate of 11.99\%. The Mid layers have produced validation rates ranging from approximately 24.49\% to 61.83\%, with a standard deviation of 10.38. Mid-deep layers range from 14.43\% to 54.02\%, with a standard deviation of 12.89, whereas Deep layers range from 4.41\% to 29.07\%, with a standard deviation of 5.65.

\begin{figure}[!ht]
    \centering
    \includegraphics[width=\textwidth]{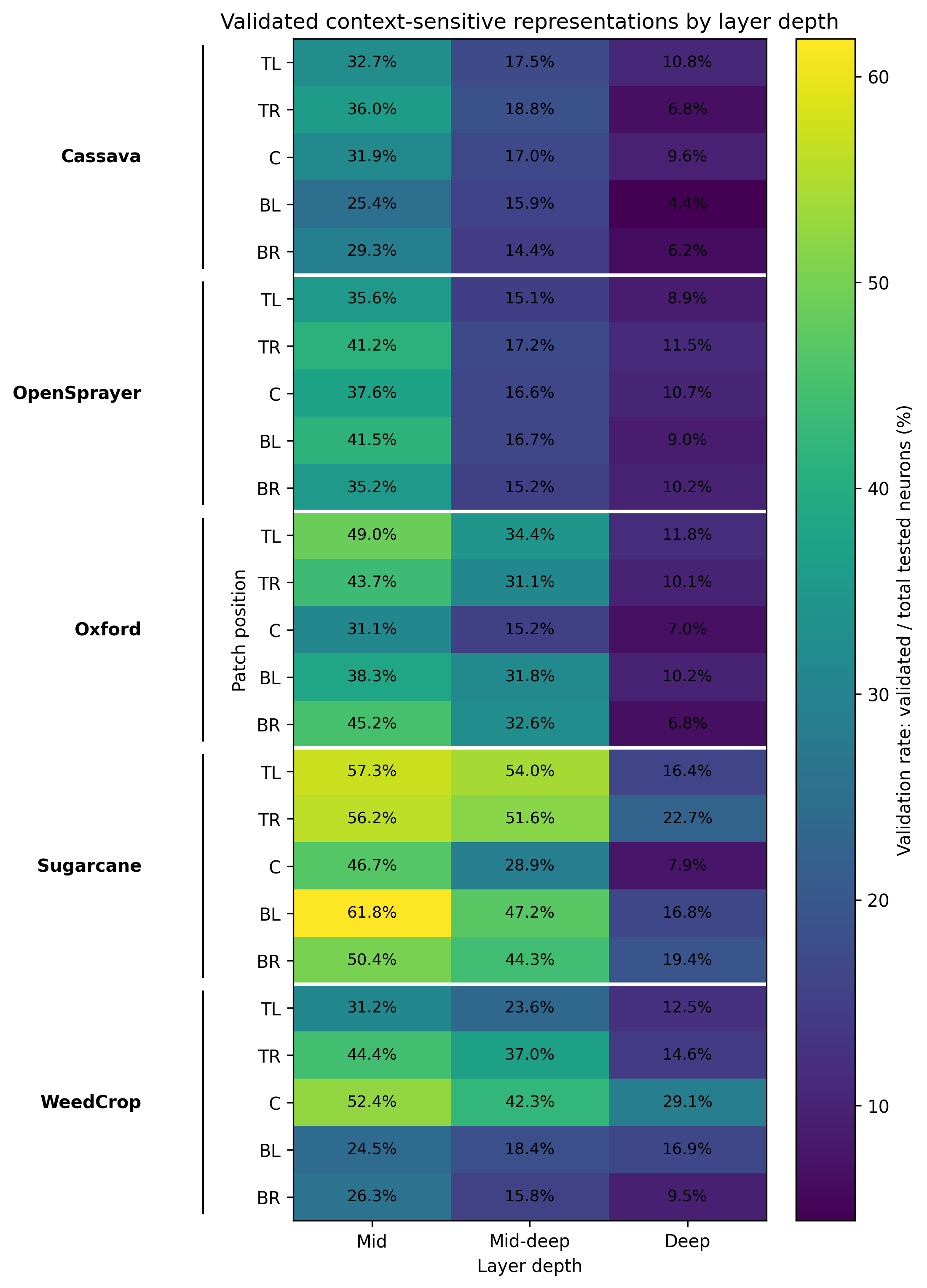}
    \caption{Validation rates of context-sensitive neuron-feature pairs across different layers for each dataset-patch configuration.}
    \label{fig:vr_dept}
\end{figure}

The layer-depth heatmap in Figure~\ref{fig:vr_dept} further shows how this pattern varies across individual datasets and patch locations. In most configurations, the Mid layer has produced the highest validation rate, while the Deep layer has produced the lowest. This tendency is especially pronounced for Sugarcane, where Mid-layer rates reach 61.8\% for the bottom-left patch and remain above 46\% for all patch locations. In contrast, corresponding Deep-layer rates remain substantially lower. WeedCrop provides another notable case: the center patch reaches 52.4\% at the Mid depth and decreases to 42.3\% at Mid-deep and 29.1\% at Deep.

At the same time, the heatmap indicates that the magnitude of the decrease is dataset-dependent. Cassava and OpenSprayer display comparatively low Deep-layer rates across all spatial regions, whereas Sugarcane and WeedCrop retain somewhat stronger validation at deeper levels in selected patches. Nevertheless, the overall ordering remains strongly weighted toward the earlier analyzed depth.
These results indicate that semantically interpretable associations with contextual patch properties are most strongly represented at the Mid depth, become less prevalent at Mid-deep layers, and are least frequently validated at the Deep depth. This pattern suggests that contextual visual characteristics are encoded most prominently during intermediate stages of hierarchical feature formation, while deeper representations become increasingly selective and task-oriented. Thus, RQ1 is answered by the consistent concentration of validated context-sensitive neuron representations in Mid-level CNN layers, followed by a progressive reduction through Mid-deep and Deep layers.

\begin{figure}[!ht]
    \centering
    \includegraphics[width=\textwidth]{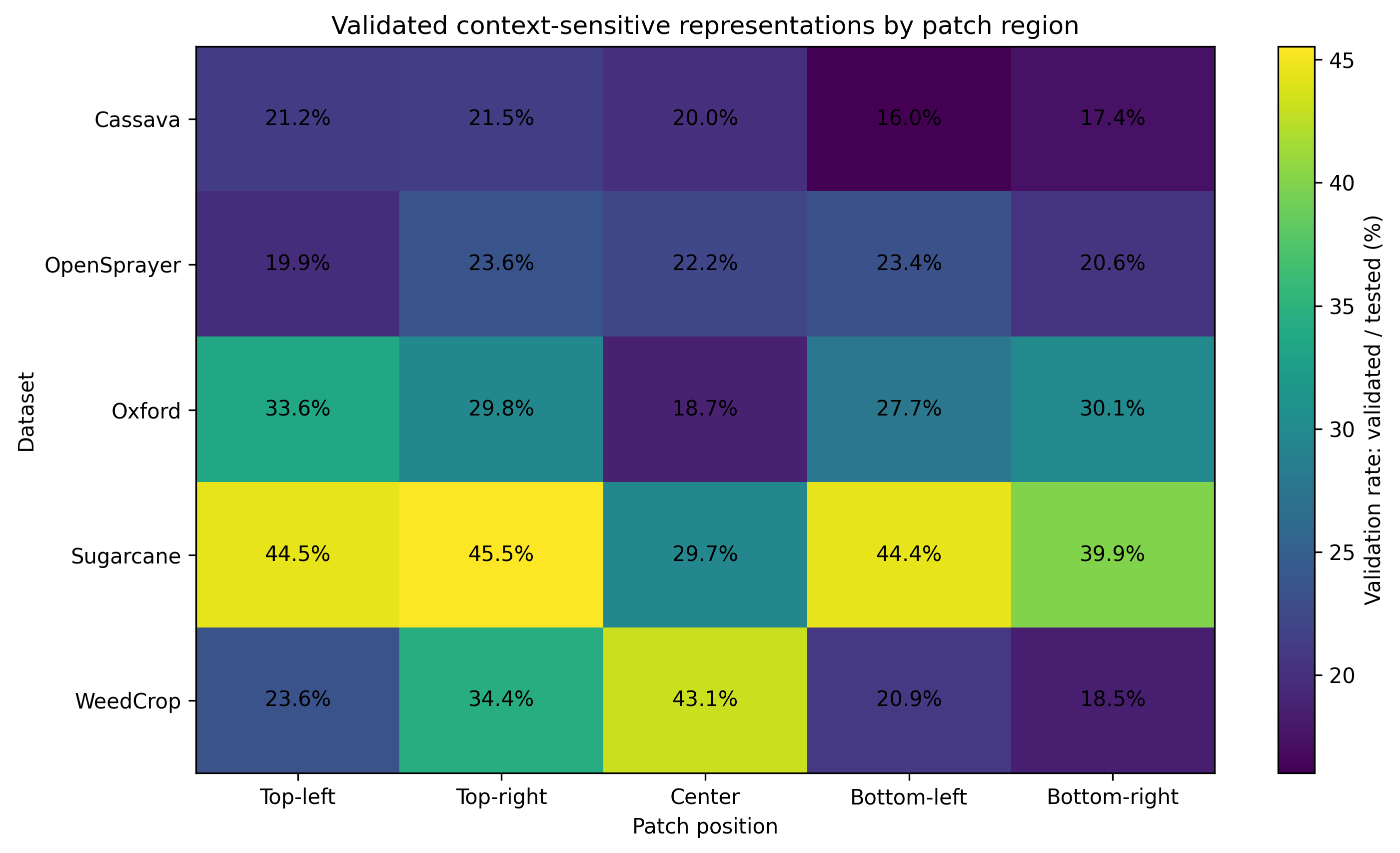}
    \caption{Dataset-wise validation rates of context-sensitive neuron-feature representations across the five predefined spatial patch locations.}
    \label{fig:vr_dept_patch}
\end{figure}

Furthermore, Figure~\ref{fig:vr_dept_patch} reveals substantial variation across datasets rather than a single universally dominant spatial region. For Cassava, validation rates remain relatively similar across the top-left, top-right, and center patches, ranging from 20.0\% to 21.5\%, while the bottom-left and bottom-right regions show lower rates of 16.0\% and 17.4\%, respectively. OpenSprayer exhibits a comparatively uniform distribution, with rates ranging from 19.9\% to 23.6\% across the five patch locations.

Oxford exhibits stronger spatial variation. The top-left patch reaches 33.6\%, followed by the bottom-right at 30.1\% and top-right at 29.8\%, whereas the center patch decreases to 18.7\%. Sugarcane has produced the highest overall patch-wise validation levels, with the top-right, top-left, and bottom-left regions reaching approximately 45.5\%, 44.5\%, and 44.4\%, respectively. In contrast, WeedCrop demonstrates a different spatial pattern, where the center patch reaches 43.1\%, substantially exceeding the top-left, bottom-left, and bottom-right regions. To examine whether any patch location is consistently dominant across datasets, the validation rates across all 15 dataset-depth observations associated with each patch position are also summarized in Table~\ref{tab:patch_validation}. The mean patch-wise validation rates are approximately 29.53\% for top-right, 27.39\% for top-left, 25.61\% for center, 25.25\% for bottom-left, and 24.06\% for bottom-right. The relatively close overall means, combined with the dataset-specific patterns visible in Figure~\ref{fig:vr_dept_patch}, indicate that no single spatial patch location consistently dominates across all datasets. Although the top-right and top-left regions exhibit modestly higher overall mean validation rates, the strongest spatial region varies by dataset. Sugarcane shows high validation across several peripheral patches, WeedCrop is dominated by the center region, and Oxford favors the top-left region. This variation suggests that context-sensitive neuron representations are not tied to a fixed image coordinate across datasets. Instead, the spatial distribution of semantically validated representations appears to depend on the visual characteristics and contextual structure associated with each dataset.

\begin{table}[ht]
\centering
\caption{Patch-wise summary of neuron-feature validation rates across datasets and analyzed layer depths.}
\label{tab:patch_validation}
\resizebox{\textwidth}{!}{%
\begin{tabular}{|C{3cm}|C{2.3cm}|C{4.5cm}|C{1cm}|C{1.5cm}|C{1.8cm}|}
\hline
\textbf{Patch position} &
\textbf{Observations} &
\textbf{Mean validation rate (\%)} &
\textbf{SD} &
\textbf{Min (\%)} &
\textbf{Max (\%)} \\ \hline

Top-left     & 15 & 27.39 & 16.25 & 8.87 & 57.32 \\ \hline
Top-right    & 15 & 29.53 & 16.08 & 6.77 & 56.20 \\ \hline
Center       & 15 & 25.61 & 14.85 & 6.96 & 52.43 \\ \hline
Bottom-left  & 15 & 25.25 & 15.97 & 4.41 & 61.83 \\ \hline
Bottom-right & 15 & 24.06 & 14.77 & 6.23 & 50.37 \\ \hline
\end{tabular}
}
\end{table}

Having established that validated context-sensitive representations are concentrated primarily at intermediate depths but vary spatially across datasets, the next analysis examines which visual feature families most consistently characterize these representations.

\subsection{RQ2 - Which Feature Families Dominate?}

RQ2 examines which visual feature families most consistently dominates in the hidden-neuron representations associated with contextual image patches. For this analysis, the statistically tested neuron-feature pairs are grouped by feature family across the analyzed datasets, patch locations, architectures, and layer depths. Within each feature family, the validation rate is calculated. 

\begin{figure}[!ht]
    \centering
    \includegraphics[width=\textwidth]{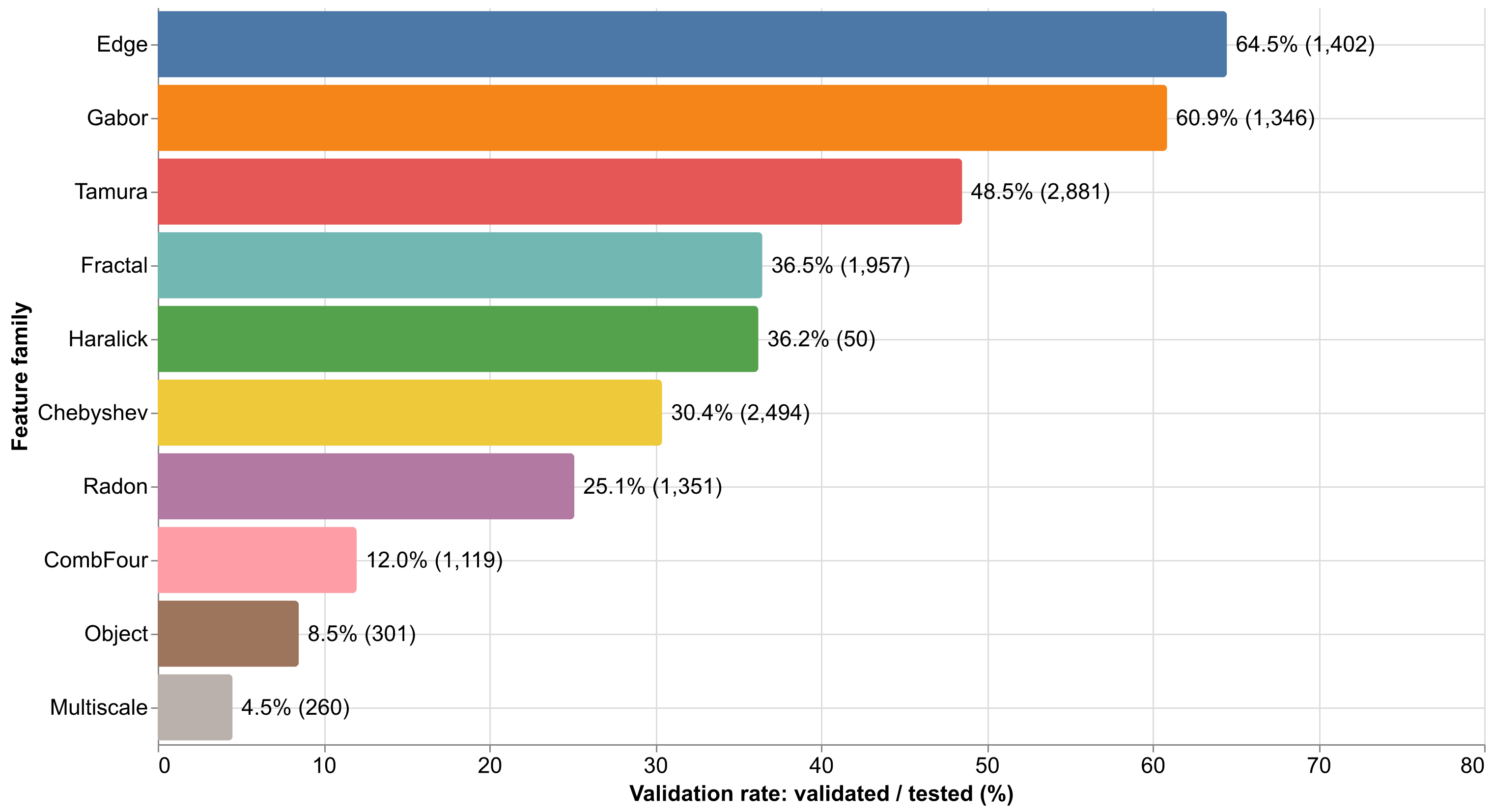}
    \caption{Overall validation outcomes for the analyzed UDAT feature families.}
    \label{fig:feature_family_bar}
\end{figure}

The overall analysis reveals a clear dominance of several edge- and texture-related descriptors. As shown in Figure~\ref{fig:feature_family_bar}, Edge features achieve the highest validation rate at 64.5\%, with 1,402 validated pairs among 2,174 statistically tested pairs. Gabor features follow closely at 60.88\% with 1,346 of 2,211 tested pairs validated, while Tamura features reach 48.52\%, corresponding to 2,881 validated pairs among 5,938 tested cases. These three feature families therefore constitute the strongest overall associations with the validated hidden-neuron representations.

\begin{figure}[!ht]
    \centering
    \includegraphics[width=0.96\textwidth]{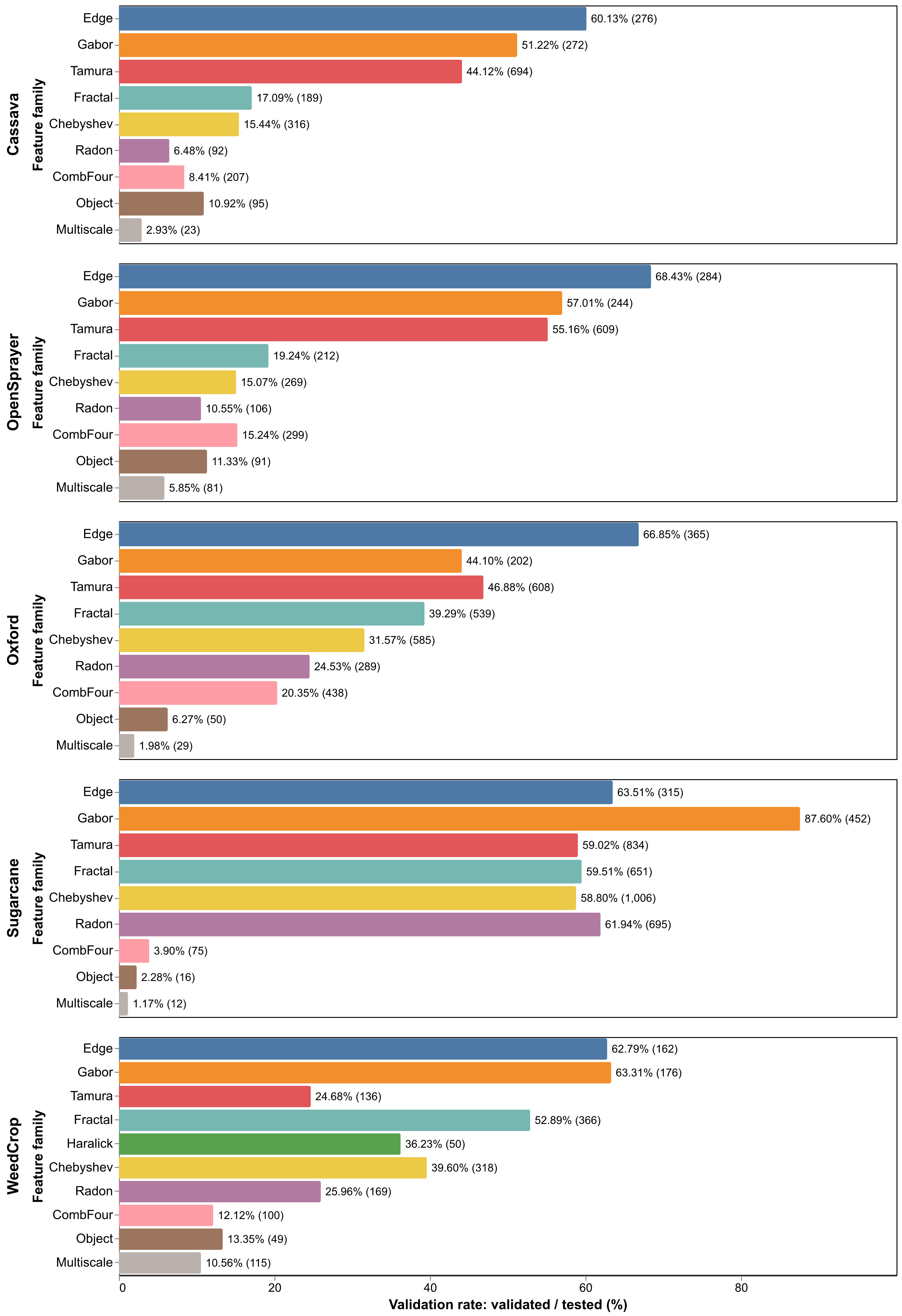}
    \caption{Feature-family validation rates grouped by dataset across the analyzed patch locations, architectures, and layer depths.}
    \label{fig:feature_family_datawise}
\end{figure}

\begin{figure}[!ht]
    \centering
    \includegraphics[width=0.96\textwidth]{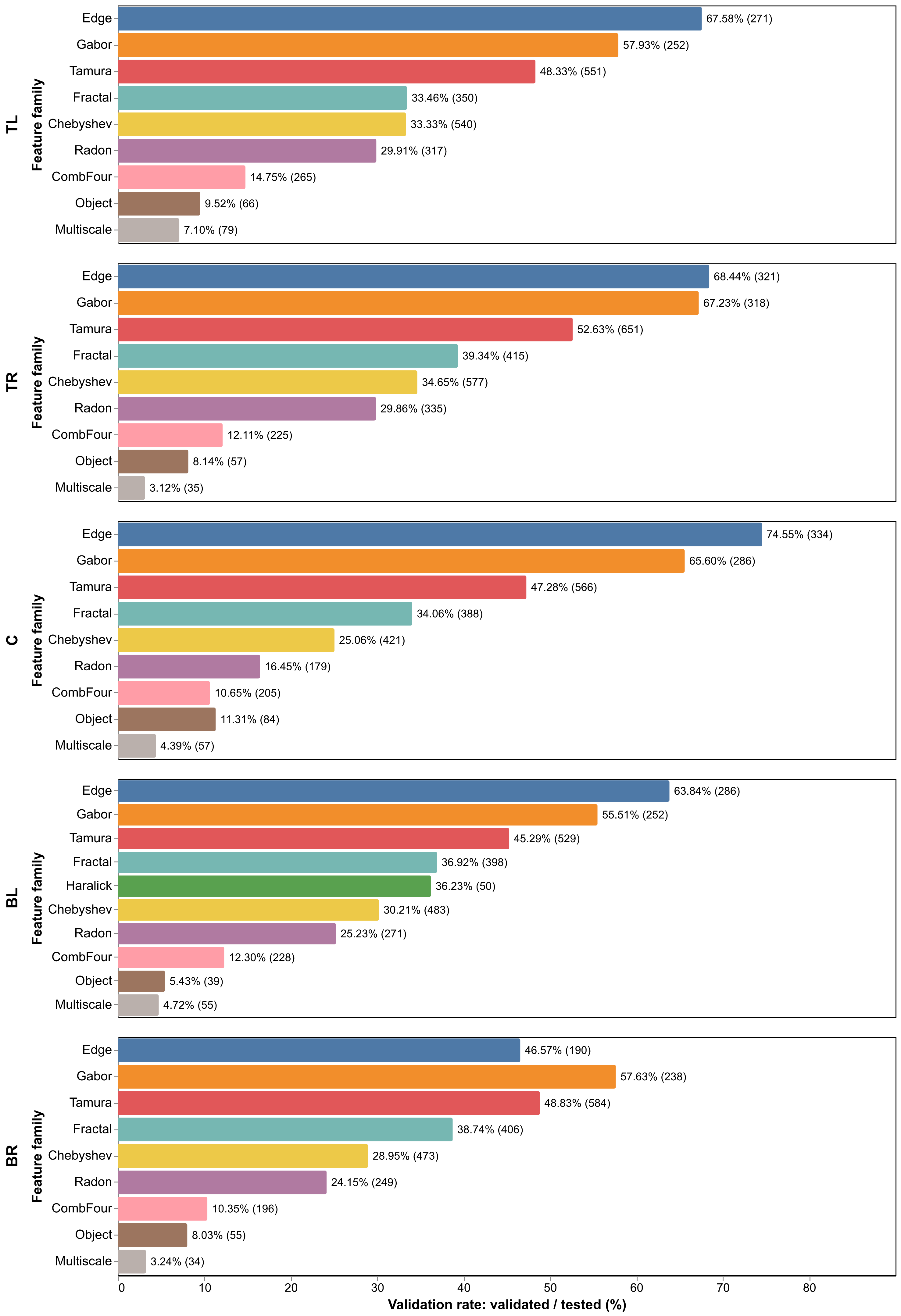}
    \caption{Feature-family validation rates grouped by spatial patch position across the analyzed datasets, architectures, and layer depths.}
    \label{fig:feature_family_patchwise}
\end{figure}

A second group of feature families exhibits more moderate validation rates. Fractal features reach 36.47\%, followed by Chebyshev at 30.42\% and Radon at 25.13\%. In contrast, CombFour, Object, and Multiscale features show substantially lower validation rates of 12.00\%, 8.51\%, and 4.52\%, respectively. Haralick features show a validation rate of 36.23\%; however, this result is based on only 138 statistically tested pairs from one dataset and one patch location and is therefore can be interpreted separately from the feature families represented across all five datasets and five patch positions.
The prominence of Edge, Gabor, and Tamura features indicates that the validated context-sensitive neuron representations are most frequently associated with visual characteristics related to edge organization, directional or frequency-sensitive texture, and perceptual texture structure. Importantly, these results are interpreted at the level of the neuron-feature pair rather than the neuron alone. An individual neuron may therefore be associated with more than one validated feature family. To determine whether the aggregate ranking is driven by only a subset of experimental conditions, the same validation-rate calculation is repeated after grouping the tested neuron-feature pairs separately by dataset and by patch position. The resulting comparisons are shown in Figures~\ref{fig:feature_family_datawise} and ~\ref{fig:feature_family_patchwise}.

Across datasets, Edge and Gabor features remain among the strongest families in nearly every case. Edge validation rates range from 60.13\% for Cassava to 68.43\% for OpenSprayer, while Gabor ranges from 44.10\% for Oxford to 87.60\% for Sugarcane. Tamura also exhibits relatively high validation in Cassava (44.12\%), OpenSprayer (55.16\%), Oxford (46.88\%), and Sugarcane (59.02\%), although its rate decreases to 24.68\% for WeedCrop.

The remaining feature families show greater dataset dependence. Fractal features are relatively weak for Cassava (17.09\%) and OpenSprayer (19.24\%) but become substantially stronger for Oxford (39.29\%), Sugarcane (59.51\%), and WeedCrop (52.89\%). A similar pattern appears for Chebyshev and Radon, both of which reach their strongest validation in Sugarcane at 58.80\% and 61.94\%, respectively. In contrast, CombFour, Object, and Multiscale remain comparatively low across most datasets. These results indicate that Edge and Gabor provide the most stable high-validation pattern across the five datasets, whereas several secondary feature families exhibit considerably greater dataset-specific variation.
The patch-wise analysis in Figure~\ref{fig:feature_family_patchwise} reveals a similarly consistent pattern. Edge features achieve the highest validation rate in four of the five spatial regions, reaching 67.58\% in the top-left, 68.44\% in the top-right, 74.55\% in the center, and 63.84\% in the bottom-left patches. In the bottom-right region, Gabor becomes the strongest feature family at 57.63\%, compared with 46.57\% for Edge. Gabor also remains consistently strong across the other patch positions, ranging from 55.51\% to 67.23\%, while Tamura maintains rates between approximately 45.29\% and 52.63\%. Fractal, Chebyshev, and Radon show moderate but more variable behavior across regions, whereas CombFour, Object, and Multiscale remain comparatively weak.

\begin{figure}[!ht]
    \centering
    \includegraphics[width=0.95\textwidth]{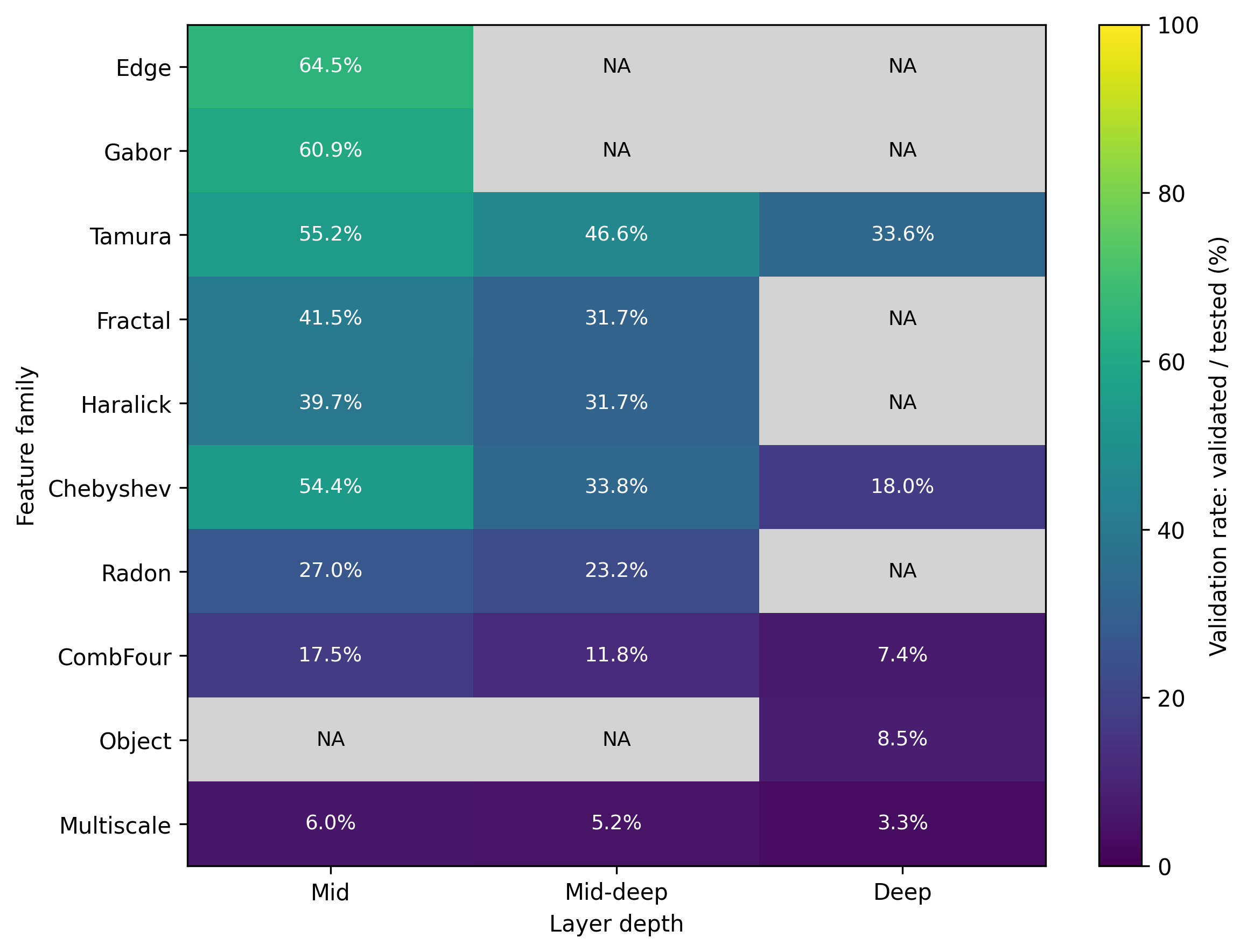}
    \caption{Validation rates of neuron-feature pairs across feature families and CNN layer depths.}
    \label{fig:feature_family_depth}
\end{figure}

Taken together, the dataset-wise and patch-wise analyses show that the high overall validation rates of Edge and Gabor are not attributable to a single dataset or spatial location. Instead, both families remain consistently prominent across heterogeneous experimental settings, with Tamura providing an additional but somewhat more variable texture-related signal.
Furthermore, to assess how feature-family associations vary with CNN depth, validation rates are computed for each layer-feature combination as validated pairs divided by statistically tested pairs and reported in Figure~\ref{fig:feature_family_depth}. Combinations with no statistically tested pairs are reported as NA. 

In the heatmap we can see that several of the strongest feature families are concentrated at the Mid depth. Edge and Gabor are represented only at the Mid depth in the tested data and achieve validation rates of 64.5\% and 60.9\%, respectively. Tamura is represented across all three depths and exhibits a progressive decline from 55.2\% at Mid to 46.6\% at Mid-deep and 33.6\% at Deep. A similar depth-related decrease is observed for Chebyshev, from 54.4\% at Mid to 33.8\% at Mid-deep and 18.0\% at Deep.
Fractal and Haralick are represented at Mid and Mid-deep depths and also show higher validation at Mid, decreasing from 41.5\% to 31.7\% for Fractal and from 39.7\% to 31.7\% for Haralick. Radon shows a smaller reduction from 27.0\% to 23.2\%, while CombFour decreases from 17.5\% at Mid to 11.8\% at Mid-deep and 7.4\% at Deep. Multiscale remains consistently low across all depths, ranging from 3.3\% to 6.0\%, whereas Object features are represented only at the Deep depth and achieve an 8.5\% validation rate.

Overall, the rate-based comparison reinforces the depth pattern identified in RQ1. For feature families represented across multiple depths, validation is generally strongest at the Mid level and decreases toward deeper layers. This suggests that the strongest semantic associations with contextual visual properties are concentrated primarily in intermediate CNN representations, although the degree of decline varies by feature family. Having identified both where context-sensitive representations emerge and which visual feature families characterize them, the next analysis examines how these semantic validation patterns relate to the degree of context bias observed in the prior patch-based study.

\subsection{RQ3 - Relationship Between Context Bias and Semantic Concept Emergence}

RQ3 examines how the emergence of validated semantic concepts relates to the degree of context bias observed across the analyzed datasets. Table~\ref{tab:dataset_bias_validation} represents the neuron-feature validation statistics on Dataset-level bias and above chance patch accuracy. The above-chance patch accuracy represents the model's behavioral dependence on contextual image regions, whereas the dataset-level validation rate represents the proportion of statistically tested neuron-feature associations that can be semantically characterized and satisfy the final validation criterion.
Sugarcane exhibits the highest semantic validation rate at 40.55\%, despite having only moderate context bias. In contrast, Cassava and OpenSprayer exhibit high context bias but considerably lower validation rates of 19.23\% and 21.97\%, respectively. WeedCrop also exhibits high context bias, yet its validation rate reaches 29.03\%, remaining below that of Sugarcane. Oxford provides another contrasting case, achieving a validation rate of 27.95\% despite showing only minimal above-chance patch performance.
\begin{table}[!ht]
\centering
\caption{Dataset-level bias and neuron-feature validation statistics.}
\vspace{0.1cm}
\label{tab:dataset_bias_validation}
\resizebox{\textwidth}{!}{%
\begin{tabular}{
|C{2.2cm}
|C{1.8cm}
|C{2.2cm}
|C{1.8cm}
|C{2.4cm}
|C{1.4cm}
|C{1.5cm}
|C{1.8cm}|
}
\hline
Dataset &
Bias category &
Mean patch accuracy &
Random chance &
Above-chance patch accuracy &
Tested &
Validated &
Validation rate \\
\hline
Cassava     & High     & 33.62\% & 20.00\% & 13.62\% & 11,251 & 2,164 & 19.23\% \\
\hline
OpenSprayer & High     & 70.18\% & 50.00\% & 20.18\% & 9,989  & 2,195 & 21.97\% \\
\hline
Oxford      & Low      & 1.70\%  & 0.98\%  & 0.72\%  & 11,115 & 3,107 & 27.95\% \\
\hline
Sugarcane   & Moderate & 16.92\% & 9.1\%  & 7.83\%  & 10,006 & 4,057 & 40.55\% \\
\hline
WeedCrop    & High     & 91.06\% & 50.00\% & 41.06\% & 5,652  & 1,641 & 29.03\% \\
\hline
\end{tabular}
}
\end{table}

The dataset rankings differ between context-bias severity and semantic concept emergence. Stronger dependence on contextual regions does not necessarily lead to more semantically validated neuron-feature associations. Similarly, datasets with moderate or low context bias can still show relatively high semantic concept emergence. Overall, RQ3 shows that stronger context bias does not necessarily correspond to greater semantic concept emergence in the analyzed hidden representations.

These results indicate that behavioral context dependence and semantic concept emergence capture distinct aspects of CNN behavior. High patch-based classification performance reflects the extent to which a model can exploit information contained within restricted image regions, whereas the validation rate reflects the extent to which hidden-neuron activations can be associated with the visual feature concepts represented in the constructed knowledge graph. Consequently, a dataset may exhibit strong contextual dependence without producing a proportionally larger number of semantically validated neuron–feature associations.

\section{Conclusion}
\label{conclude}

Previous studies have demonstrated that CNNs can achieve above-chance classification accuracy using image patches containing little or no task-relevant content, revealing potential dependence on contextual information. Building on these findings, this study investigates how such information is represented within CNN hidden layers. We employ a neuro-symbolic approach that combines hidden-neuron activation analysis, UDAT image descriptors, OWL knowledge graphs, and Description Logic Concept Induction to identify interpretable associations between neuron activations and visual characteristics of contextual image patches. Across five datasets, five patch locations, four CNN architectures, and three network depths, the analysis evaluates 105,000 neuron-feature-family combinations, of which 77,027 produce candidate concepts, 48,013 undergo statistical testing, and 13,164 satisfy the validation criteria.

The findings reveal several patterns in the internal representations of CNNs. Mid layers exhibit the highest mean validation rate (40.2\%), followed by mid-deep (26.9\%) and deep layers (12.0\%), indicating that the evaluated image descriptors characterize mid-layer activations more consistently. Spatial analysis reveals no universally dominant patch location, as the regions producing the highest validation rates vary across datasets. Among the evaluated feature families, Edge, Gabor, and Tamura achieve the strongest and most consistent validation results across datasets and patch locations, suggesting that edge structure and texture-related characteristics are particularly prominent in the representations associated with contextual image patches. However, the degree of concept validation does not directly correspond to the severity of context bias observed in the previous patch-based study. Sugarcane, which exhibits moderate context bias, achieves the highest overall validation rate (40.5\%), whereas highly biased Cassava achieves the lowest (19.2\%).

Together, these findings extend patch-based context-bias analysis from observing classification behavior to examining the visual characteristics associated with hidden-neuron activations. They also suggest that strong contextual dependence does not necessarily imply a greater proportion of statistically validated neuron-feature associations. 

Several limitations affect how far these findings reach. The feature families were assigned to layers by an LLM-guided assessment rather than crossed with all layers. Edge and Gabor were evaluated only at mid layers, so their high validation rates cannot be separated from the depth effect, and only Tamura, among the strongest families, was tested at all three depths. Zernike moments yielded no testable concepts, and Haralick features were tested on a single dataset and patch location. While the proposed analysis provides interpretable evidence of internal representations, these associations alone do not establish which features causally influence classification decisions.  
Future work can investigate this relationship through controlled feature interventions and examine whether modifying or removing identified contextual characteristics changes model predictions, thereby supporting a more direct assessment of their contribution to context bias.

\section{Code and Data Availability}

The code and data associated with this study will be made publicly available upon publication at \url{https://github.com/Samatha1994/XAI_Understanding_Bias_Project/}.

\appendix

\section{LLM-Guided Feature-Layer Assignment}
\label{app:llm-assessment}

The feature-layer assessment described in Section~\ref{sec:udat} was 
conducted using Claude Sonnet (Anthropic, 2026). For each architecture, the model was provided with the architecture name, the selected layer names, and the 13 UDAT feature family names, and queried as follows:

\begin{quote}
\textit{For the following layers of [ARCHITECTURE]: [LAYER NAMES]  
and the following feature families: Edge Statistics, Combined First 
Four Moments, Chebyshev-Fourier Transform, Multiscale Histogram, 
Tamura Texture, Haralick Texture, Gabor Filters, Zernike Moments, 
Object Statistics, Radon Transform, Color Histogram, Fractal Features, 
and Intensity Statistics --- which features should be tested for each layer, and which should be skipped?}
\end{quote}

This prompt was issued separately for each of the four architectures, 
with layer names substituted accordingly. The model returned a 
classification of each layer-feature-family combination as 
\emph{test}, \emph{optional/borderline}, or \emph{skip}, along with 
theoretical justification for each decision based on known properties of convolutional feature hierarchies.

\bibliographystyle{plain}
\bibliography{sample}

@article{plantvillage,
title={Using deep learning for image-based plant disease detection},
volume={7},
DOI={10.3389/fpls.2016.01419},
journal={Frontiers in Plant Science},
author={Mohanty, Sharada P. and Hughes, David P. and Salathé, Marcel},
year={2016},
month={Sep}}

@incollection{plantdoc,
  title={PlantDoc: A dataset for visual plant disease detection},
  author={Singh, Davinder and Jain, Naman and Jain, Pranjali and Kayal, Pratik and Kumawat, Sudhakar and Batra, Nipun},
  booktitle={Proceedings of the 7th ACM IKDD CoDS and 25th COMAD},
  publisher = {Association for Computing Machinery},
  pages={249--253},
  year={2020}
}

@article{DeepWeeds2019,
  author = {Alex Olsen and
    Dmitry A. Konovalov and
    Bronson Philippa and
    Peter Ridd and
    Jake C. Wood and
    Jamie Johns and
    Wesley Banks and
    Benjamin Girgenti and
    Owen Kenny and 
    James Whinney and
    Brendan Calvert and
    Mostafa {Rahimi Azghadi} and
    Ronald D. White},
  title = {DeepWeeds: A Multiclass Weed Species Image Dataset for Deep Learning},
  journal = {Scientific Reports},
  year = 2019,
  number = 2058,
  month = 2,
  volume = 9,
  issue = 1,
  day = 14,
  url = "https://doi.org/10.1038/s41598-018-38343-3",
  doi = "10.1038/s41598-018-38343-3"
}

@misc{cassava,
    author = {ErnestMwebaze and Jesse Mostipak and Joyce and Julia Elliott and Sohier Dane},
    title = {Cassava Leaf Disease Classification},
    year = {2020},
    howpublished = {\url{https://kaggle.com/competitions/cassava-leaf-disease-classification}},
    note = {Kaggle}
}

@article{weedCrop,
  title={Dataset of annotated food crops and weed images for robotic computer vision control},
  author={Sudars, Kaspars and Jasko, Janis and Namatevs, Ivars and Ozola, Liva and Badaukis, Niks},
  journal={Data in brief},
  volume={31},
  pages={105833},
  year={2020},
  publisher={Elsevier}
}

@misc{openSprayer,
  title={Open Sprayer Images},
  author={Armstrong, Gavin},
  year={2023},
  publisher={Kaggle},
  url={https://www.kaggle.com/datasets/gavinarmstrong/open-sprayer-images},
  note={Accessed: June 2025}
}

@article{sugarcane,
  title={Sugarcane leaf dataset: A dataset for disease detection and classification for machine learning applications},
  author={Thite, Sandip and Suryawanshi, Yogesh and Patil, Kailas and Chumchu, Prawit},
  journal={Data in Brief},
  volume={53},
  pages={110268},
  year={2024},
  publisher={Elsevier}
}

@INPROCEEDINGS{oxford102,
  author={Nilsback, Maria-Elena and Zisserman, Andrew},
  booktitle={2008 Sixth Indian Conference on Computer Vision, Graphics \& Image Processing}, 
  title={Automated Flower Classification over a Large Number of Classes}, 
  year={2008},
  volume={},
  number={},
  pages={722-729},
  doi={10.1109/ICVGIP.2008.47}}

@article{shamir2017udat,
  title={Udat: A multi-purpose data analysis tool},
  author={Shamir, Lior},
  journal={Astrophysics Source Code Library},
  pages={ascl--1704},
  year={2017}
}

@inproceedings{dalal2024value,
  title={On the value of labeled data and symbolic methods for hidden neuron activation analysis},
  author={Dalal, Abhilekha and Rayan, Rushrukh and Barua, Adrita and Vasserman, Eugene Y and Sarker, Md Kamruzzaman and Hitzler, Pascal},
  booktitle={International Conference on Neural-Symbolic Learning and Reasoning},
  pages={109--131},
  year={2024},
  organization={Springer}
}

@article{DBLP:journals/ml/LehmannH10,
  author    = {Jens Lehmann and Pascal Hitzler},
  title     = {Concept learning in description logics using refinement operators},
  journal   = {Mach. Learn.},
  volume    = {78},
  number    = {1-2},
  pages     = {203--250},
  year      = {2010},
  doi       = {10.1007/s10994-009-5146-2},
  bibsource = {dblp computer science bibliography, https://dblp.org}
}

@book{DBLP:books/crc/Hitzler2010,
  author    = {Pascal Hitzler and
               Markus Kr{\"{o}}tzsch and
               Sebastian Rudolph},
  title     = {Foundations of Semantic Web Technologies},
  publisher = {Chapman and Hall/CRC Press},
  year      = {2010},
  url       = {http://www.semantic-web-book.org/},
  isbn      = {9781420090505},
  bibsource = {dblp computer science bibliography, https://dblp.org}
}

@article{DBLP:journals/cacm/Hitzler21,
  author    = {Pascal Hitzler},
  title     = {A review of the semantic web field},
  journal   = {Commun. {ACM}},
  volume    = {64},
  number    = {2},
  pages     = {76--83},
  year      = {2021},
  url       = {https://doi.org/10.1145/3397512},
  doi       = {10.1145/3397512},
  bibsource = {dblp computer science bibliography, https://dblp.org}
}

@article{lehmann2011celoe,
  title={Class expression learning for ontology engineering},
  author={Lehmann, Jens and Auer, S{\"o}ren and B{\"u}hmann, Lorenz and Tramp, Sebastian},
  journal={Journal of Web Semantics},
  volume={9},
  number={1},
  pages={71--81},
  year={2011},
  publisher={Elsevier},
  doi={10.1016/j.websem.2011.01.001}
}

@article{dalal2026evaluating,
  title={Evaluating Contextual Bias in CNN Image Classification: Evidence from Agricultural Benchmark Datasets},
  author={Dalal, Abhilekha and Okonoda, Michael and Martinez, Eder and Shamir, Lior},
  journal={arXiv preprint arXiv:2609.14654},
  year={2026},
  doi={10.48550/arXiv.2609.14654},
  url={https://arxiv.org/abs/2609.14654}
}

@article{krizhevsky2012imagenet,
  title={ImageNet Classification with Deep Convolutional Neural Networks},
  author={Krizhevsky, Alex and Sutskever, Ilya and Hinton, Geoffrey E.},
  journal={Advances in Neural Information Processing Systems},
  volume={25},
  pages={1097--1105},
  year={2012}
}

@inproceedings{he2016deep,
  title={Deep Residual Learning for Image Recognition},
  author={He, Kaiming and Zhang, Xiangyu and Ren, Shaoqing and Sun, Jian},
  booktitle={Proceedings of the IEEE Conference on Computer Vision and Pattern Recognition (CVPR)},
  pages={770--778},
  year={2016}
}

@article{geirhos2020shortcut,
  title={Shortcut Learning in Deep Neural Networks},
  author={Geirhos, Robert and Jacobsen, J{\"o}rn-Henrik and Michaelis, Claudio and Zemel, Richard and Brendel, Wieland and Bethge, Matthias and Wichmann, Felix A.},
  journal={Nature Machine Intelligence},
  volume={2},
  number={11},
  pages={665--673},
  year={2020},
  publisher={Nature Publishing Group}
}

@inproceedings{ilyas2019adversarial,
  title={Adversarial Examples Are Not Bugs, They Are Features},
  author={Ilyas, Andrew and Santurkar, Shibani and Tsipras, Dimitris and Engstrom, Logan and Tran, Brandon and Madry, Aleksander},
  booktitle={Advances in Neural Information Processing Systems (NeurIPS)},
  volume={32},
  year={2019}
}

@article{shahroudnejad2021survey,
  title={A Survey on Understanding, Visualizations, and Explanation of Deep Neural Networks},
  author={Shahroudnejad, Atefeh},
  journal={arXiv preprint arXiv:2102.01792},
  year={2021}
}

@inproceedings{beery2018recognition,
  title={Recognition in Terra Incognita},
  author={Beery, Sara and Van Horn, Grant and Perona, Pietro},
  booktitle={Proceedings of the European Conference on Computer Vision (ECCV)},
  pages={456--473},
  year={2018}
}

@article{xiao2020noise,
  title={Noise or Signal: The Role of Image Backgrounds in Object Recognition},
  author={Xiao, Kai and Engstrom, Logan and Ilyas, Andrew and Madry, Aleksander},
  journal={arXiv preprint arXiv:2006.09994},
  year={2020}
}

@article{ross2024shortcut,
  title={Shortcut learning in medical AI hinders generalization: method for estimating AI model generalization without external data},
  author={Ly, Cathy Ong and Unnikrishnan, Balagopal and Tadic, Tony and Patel, Tirth and Duhamel, Joe and Kandel, Sonja and Moayedi, Yasbanoo and Brudno, Michael and Hope, Andrew and Ross, Heather and McIntosh, Chris},
  journal={npj Digital Medicine},
  volume={7},
  number={1},
  pages={124},
  year={2024},
  publisher={Nature Publishing Group UK London}
}

@inproceedings{aniraj2023masking,
  title={Masking Strategies for Background Bias Removal in Computer Vision Models},
  author={Aniraj, Ananthu and Dantas, Cassio F. and Ienco, Dino and Marcos, Diego},
  booktitle={Proceedings of the IEEE/CVF International Conference on Computer Vision (ICCV) Workshops},
  pages={4454--4463},
  year={2023}
}

@inproceedings{shetty2019not,
  title={Not Using the Car to See the Sidewalk--Quantifying and Controlling the Effects of Context in Classification and Segmentation},
  author={Shetty, Rakshith and Schiele, Bernt and Fritz, Mario},
  booktitle={Proceedings of the IEEE/CVF conference on computer vision and pattern recognition},
  pages={8218--8226},
  year={2019}
}

@inproceedings{ribeiro2016should,
  title={" Why should i trust you?" Explaining the predictions of any classifier},
  author={Ribeiro, Marco Tulio and Singh, Sameer and Guestrin, Carlos},
  booktitle={Proceedings of the 22nd ACM SIGKDD international conference on knowledge discovery and data mining},
  pages={1135--1144},
  year={2016}
}

@article{zech2018variable,
  title={Variable generalization performance of a deep learning model to detect pneumonia in chest radiographs: a cross-sectional study},
  author={Zech, John R and Badgeley, Marcus A and Liu, Manway and Costa, Anthony B and Titano, Joseph J and Oermann, Eric Karl},
  journal={PLoS medicine},
  volume={15},
  number={11},
  pages={e1002683},
  year={2018},
  publisher={Public Library of Science}
}

@article{martinez2026detection,
  title={Detection and profiling of dataset bias in CNN analysis of images},
  author={Martinez, Eder and Okonoda, Michael and Dalal, Abhilekha and Shamir, Lior},
  journal={Computing in Science \& Engineering},
  year={2026},
  publisher={IEEE}
}

@article{shamir2008evaluation,
  title={Evaluation of face datasets as tools for assessing the performance of face recognition methods},
  author={Shamir, Lior},
  journal={International Journal of Computer Vision},
  volume={79},
  number={3},
  pages={225--230},
  year={2008},
  publisher={Springer}
}

@inproceedings{zhou2016learning,
  title={Learning Deep Features for Discriminative Localization},
  author={Zhou, Bolei and Khosla, Aditya and Lapedriza, Agata and Oliva, Aude and Torralba, Antonio},
  booktitle={Proceedings of the IEEE Conference on Computer Vision and Pattern Recognition (CVPR)},
  pages={2921--2929},
  year={2016}
}

@INPROCEEDINGS{gradcam_batra,
  author={Selvaraju, Ramprasaath R. and Cogswell, Michael and Das, Abhishek and Vedantam, Ramakrishna and Parikh, Devi and Batra, Dhruv},
  booktitle={2017 IEEE International Conference on Computer Vision (ICCV)}, 
  title={Grad-CAM: Visual Explanations from Deep Networks via Gradient-Based Localization}, 
  year={2017},
  volume={},
  number={},
  pages={618-626},
  doi={10.1109/ICCV.2017.74}}

@inproceedings{oikarinen2023clip,
  title={CLIP-Dissect: Automatic Description of Neuron Representations in Deep Vision Networks},
  author={Oikarinen, Tuomas and Weng, Tsui-Wei},
  booktitle={International Conference on Learning Representations (ICLR)},
  year={2023}
}

@book{dalal2024understanding,
  title={Understanding Hidden Neuron Activations Using Structured Background Knowledge and Deductive Reasoning},
  author={Dalal, Abhilekha},
  year={2024},
  publisher={Kansas State University}
}

@inproceedings{torralba2011unbiased,
  author    = {Torralba, Antonio and Efros, Alexei A.},
  title     = {Unbiased Look at Dataset Bias},
  booktitle = {Proceedings of the IEEE Conference on Computer Vision and Pattern Recognition (CVPR)},
  pages     = {1521--1528},
  year      = {2011},
  doi       = {10.1109/CVPR.2011.5995347}
}

@article{lapuschkin2019unmasking,
  author  = {Lapuschkin, Sebastian and W{\"a}ldchen, Stephan and Binder, Alexander and Montavon, Gr{\'e}goire and Samek, Wojciech and M{\"u}ller, Klaus-Robert},
  title   = {Unmasking {Clever Hans} Predictors and Assessing What Machines Really Learn},
  journal = {Nature Communications},
  volume  = {10},
  pages   = {1096},
  year    = {2019},
  doi     = {10.1038/s41467-019-08987-4}
}

@article{degrave2021ai,
  author  = {DeGrave, Alex J. and Janizek, Joseph D. and Lee, Su-In},
  title   = {{AI} for Radiographic {COVID-19} Detection Selects Shortcuts over Signal},
  journal = {Nature Machine Intelligence},
  volume  = {3},
  pages   = {610--619},
  year    = {2021},
  doi     = {10.1038/s42256-021-00338-7}
}

@article{noyan2022uncovering,
  author  = {Noyan, Mehmet Alican},
  title   = {Uncovering Bias in the {PlantVillage} Dataset},
  journal = {arXiv preprint arXiv:2206.04374},
  year    = {2022},
  doi     = {10.48550/arXiv.2206.04374}
}

@article{schramowski2020making,
  author  = {Schramowski, Patrick and Stammer, Wolfgang and Teso, Stefano and Brugger, Anna and Herbert, Franziska and Shao, Xiaoting and Luigs, Hans-Georg and Mahlein, Anne-Katrin and Kersting, Kristian},
  title   = {Making Deep Neural Networks Right for the Right Scientific Reasons by Interacting with Their Explanations},
  journal = {Nature Machine Intelligence},
  volume  = {2},
  pages   = {476--486},
  year    = {2020},
  doi     = {10.1038/s42256-020-0212-3}
}

@article{model2015comparison,
  author  = {Model, Ian and Shamir, Lior},
  title   = {Comparison of Data Set Bias in Object Recognition Benchmarks},
  journal = {IEEE Access},
  volume  = {3},
  year    = {2015},
  doi     = {10.1109/ACCESS.2015.2491921}
}

@article{dhar2021evaluation,
  author  = {Dhar, Sanchari and Shamir, Lior},
  title   = {Evaluation of the Benchmark Datasets for Testing the Efficacy of Deep Convolutional Neural Networks},
  journal = {Visual Informatics},
  year    = {2021},
  doi     = {10.1016/j.visinf.2021.10.001}
}

@article{erukude2024identifying,
  title={Identifying bias in deep neural networks using image transforms},
  author={Erukude, Sai Teja and Joshi, Akhil and Shamir, Lior},
  journal={Computers},
  volume={13},
  number={12},
  pages={341},
  year={2024},
  publisher={MDPI}
}

@Article{okonoda2026unmasking,
AUTHOR = {Okonoda, Michael and Martinez, Eder and Dalal, Abhilekha and Shamir, Lior},
TITLE = {Unmasking Biases and Reliability Concerns in Convolutional Neural Networks Analysis of Cancer Pathology Images},
JOURNAL = {Electronics},
VOLUME = {15},
YEAR = {2026},
NUMBER = {6},
ARTICLE-NUMBER = {1182},
URL = {https://www.mdpi.com/2079-9292/15/6/1182},
ISSN = {2079-9292},
DOI = {10.3390/electronics15061182}
}

@inproceedings{zeiler2014visualizing,
  author    = {Zeiler, Matthew D. and Fergus, Rob},
  title     = {Visualizing and Understanding Convolutional Networks},
  booktitle = {European Conference on Computer Vision (ECCV)},
  pages     = {818--833},
  year      = {2014},
  doi       = {10.1007/978-3-319-10590-1_53}
}

@article{simonyan2014deep,
  author  = {Simonyan, Karen and Vedaldi, Andrea and Zisserman, Andrew},
  title   = {Deep Inside Convolutional Networks: Visualising Image Classification Models and Saliency Maps},
  journal = {arXiv preprint arXiv:1312.6034},
  year    = {2014},
  doi     = {10.48550/arXiv.1312.6034}
}

@article{yosinski2015understanding,
  author  = {Yosinski, Jason and Clune, Jeff and Nguyen, Anh and Fuchs, Thomas and Lipson, Hod},
  title   = {Understanding Neural Networks Through Deep Visualization},
  journal = {arXiv preprint arXiv:1506.06579},
  year    = {2015},
  doi     = {10.48550/arXiv.1506.06579}
}

@inproceedings{bau2017network,
  author    = {Bau, David and Zhou, Bolei and Khosla, Aditya and Oliva, Aude and Torralba, Antonio},
  title     = {Network Dissection: Quantifying Interpretability of Deep Visual Representations},
  booktitle = {Proceedings of the IEEE Conference on Computer Vision and Pattern Recognition (CVPR)},
  pages     = {6541--6549},
  year      = {2017},
  doi       = {10.1109/CVPR.2017.354}
}

@inproceedings{mu2020compositional,
  author    = {Mu, Jesse and Andreas, Jacob},
  title     = {Compositional Explanations of Neurons},
  booktitle = {Advances in Neural Information Processing Systems (NeurIPS)},
  volume    = {33},
  year      = {2020},
  doi       = {10.48550/arXiv.2006.14032}
}

@inproceedings{kim2018tcav,
  author    = {Kim, Been and Wattenberg, Martin and Gilmer, Justin and Cai, Carrie and Wexler, James and Viegas, Fernanda and Sayres, Rory},
  title     = {Interpretability Beyond Feature Attribution: Quantitative Testing with Concept Activation Vectors ({TCAV})},
  booktitle = {Proceedings of the 35th International Conference on Machine Learning (ICML)},
  pages     = {2668--2677},
  year      = {2018},
  doi       = {10.48550/arXiv.1711.11279}
}

@inproceedings{ghorbani2019ace,
  author    = {Ghorbani, Amirata and Wexler, James and Zou, James and Kim, Been},
  title     = {Towards Automatic Concept-based Explanations},
  booktitle = {Advances in Neural Information Processing Systems (NeurIPS)},
  volume    = {32},
  year      = {2019},
  doi       = {10.48550/arXiv.1902.03129}
}

@inproceedings{hernandez2022milan,
  author    = {Hernandez, Evan and Schwettmann, Sarah and Bau, David and Bagashvili, Teona and Torralba, Antonio and Andreas, Jacob},
  title     = {Natural Language Descriptions of Deep Visual Features},
  booktitle = {International Conference on Learning Representations (ICLR)},
  year      = {2022},
  doi       = {10.48550/arXiv.2201.11114}
}

@inproceedings{le2025xai,
  title={An xai-based analysis of shortcut learning in neural networks},
  author={Le, Phuong Quynh and Schl{\"o}tterer, J{\"o}rg and Seifert, Christin},
  booktitle={World Conference on Explainable Artificial Intelligence},
  pages={424--445},
  year={2025},
  organization={Springer}
}

@inproceedings{ramaswamy2023overlooked,
  author    = {Ramaswamy, Vikram V. and Kim, Sunnie S. Y. and Fong, Ruth and Russakovsky, Olga},
  title     = {Overlooked Factors in Concept-Based Explanations: Dataset Choice, Concept Learnability, and Human Capability},
  booktitle = {Proceedings of the IEEE/CVF Conference on Computer Vision and Pattern Recognition (CVPR)},
  year      = {2023},
  doi       = {10.1109/CVPR52729.2023.01052}
}

@article{hitzler2020neural,
  author  = {Hitzler, Pascal and Bianchi, Federico and Ebrahimi, Monireh and Sarker, Md Kamruzzaman},
  title   = {Neural-Symbolic Integration and the {Semantic Web}},
  journal = {Semantic Web},
  volume  = {11},
  number  = {1},
  pages   = {3--11},
  year    = {2020},
  doi     = {10.3233/SW-190368}
}

@inproceedings{desousaribeiro2021aligning,
  author    = {de Sousa Ribeiro, Manuel and Leite, Jo{\~a}o},
  title     = {Aligning Artificial Neural Networks and Ontologies towards Explainable {AI}},
  booktitle = {Proceedings of the AAAI Conference on Artificial Intelligence},
  volume    = {35},
  year      = {2021},
  doi       = {10.1609/aaai.v35i6.16626}
}

@inproceedings{sarker2019efficient,
  author    = {Sarker, Md Kamruzzaman and Hitzler, Pascal},
  title     = {Efficient Concept Induction for Description Logics},
  booktitle = {Proceedings of the AAAI Conference on Artificial Intelligence},
  volume    = {33},
  pages     = {3036--3043},
  year      = {2019},
  doi       = {10.1609/aaai.v33i01.33013036}
}

@article{lehmann2009dllearner,
  author  = {Lehmann, Jens},
  title   = {{DL-Learner}: Learning Concepts in Description Logics},
  journal = {Journal of Machine Learning Research},
  volume  = {10},
  pages   = {2639--2642},
  year    = {2009}
}

@article{buhmann2016dllearner,
  author  = {B{\"u}hmann, Lorenz and Lehmann, Jens and Westphal, Patrick},
  title   = {{DL-Learner}---A Framework for Inductive Learning on the {Semantic Web}},
  journal = {Journal of Web Semantics},
  volume  = {39},
  pages   = {15--24},
  year    = {2016},
  doi     = {10.1016/j.websem.2016.06.001}
}

@article{sarker2017explaining,
  author  = {Sarker, Md Kamruzzaman and Xie, Ning and Doran, Derek and Raymer, Michael and Hitzler, Pascal},
  title   = {Explaining Trained Neural Networks with {Semantic Web} Technologies: First Steps},
  journal = {arXiv preprint arXiv:1710.04324},
  year    = {2017},
  doi     = {10.48550/arXiv.1710.04324}
}

@article{dalal2026toward,
  title={Toward a Neurosymbolic Understanding of Hidden Neuron Activations},
  author={Dalal, Abhilekha and Rayan, Rushrukh and Barua, Adrita and Akkamahadevi, Samatha Ereshi and Das, Avishek and Widmer, Cara and Vasserman, Eugene Y and Sarker, Md Kamruzzaman and Hitzler, Pascal},
  journal={Neurosymbolic Artificial Intelligence},
  volume={2},
  pages={29498732261419302},
  year={2026},
  publisher={SAGE Publications Sage UK: London, England}
}

@inproceedings{akkamahadevi2025evaluation,
  author    = {Akkamahadevi, Samatha Ereshi and Dalal, Abhilekha and Hitzler, Pascal},
  title     = {Evaluation of Concept Induction in Explainable {AI} Using Multiple Datasets},
  booktitle = {Proceedings of the AAAI Symposium Series},
  volume    = {5},
  year      = {2025},
  doi       = {10.1609/aaaiss.v5i1.35607}
}

@article{sensarma2026case,
  title={A Case Study on Concept Induction for Neuron-Level Interpretability in CNN},
  author={Sarma, Moumita Sen and Akkamahadevi, Samatha Ereshi and Hitzler, Pascal},
  journal={arXiv preprint arXiv:2603.00197},
  year={2026}
}

@article{haralick1973textural,
  author  = {Haralick, Robert M. and Shanmugam, K. and Dinstein, Its'hak},
  title   = {Textural Features for Image Classification},
  journal = {IEEE Transactions on Systems, Man, and Cybernetics},
  volume  = {SMC-3},
  number  = {6},
  pages   = {610--621},
  year    = {1973},
  doi     = {10.1109/TSMC.1973.4309314}
}

@article{tamura1978textural,
  author  = {Tamura, Hideyuki and Mori, Shunji and Yamawaki, Takashi},
  title   = {Textural Features Corresponding to Visual Perception},
  journal = {IEEE Transactions on Systems, Man, and Cybernetics},
  volume  = {8},
  number  = {6},
  pages   = {460--473},
  year    = {1978},
  doi     = {10.1109/TSMC.1978.4309999}
}

@article{manjunath1996texture,
  author  = {Manjunath, B. S. and Ma, W. Y.},
  title   = {Texture Features for Browsing and Retrieval of Image Data},
  journal = {IEEE Transactions on Pattern Analysis and Machine Intelligence},
  volume  = {18},
  number  = {8},
  pages   = {837--842},
  year    = {1996},
  doi     = {10.1109/34.531803}
}

@article{ojala2002multiresolution,
  author  = {Ojala, Timo and Pietik{\"a}inen, Matti and M{\"a}enp{\"a}{\"a}, Topi},
  title   = {Multiresolution Gray-Scale and Rotation Invariant Texture Classification with Local Binary Patterns},
  journal = {IEEE Transactions on Pattern Analysis and Machine Intelligence},
  volume  = {24},
  number  = {7},
  pages   = {971--987},
  year    = {2002},
  doi     = {10.1109/TPAMI.2002.1017623}
}

@article{orlov2008wndcharm,
  author  = {Orlov, Nikita and Shamir, Lior and Macura, Tomasz and Johnston, Josiah and Eckley, D. Mark and Goldberg, Ilya G.},
  title   = {{WND-CHARM}: Multi-Purpose Image Classification Using Compound Image Transforms},
  journal = {Pattern Recognition Letters},
  volume  = {29},
  number  = {11},
  pages   = {1684--1693},
  year    = {2008},
  doi     = {10.1016/j.patrec.2008.04.013}
}

@article{shamir2008wndchrm,
  author  = {Shamir, Lior and Orlov, Nikita and Eckley, D. Mark and Macura, Tomasz and Johnston, Josiah and Goldberg, Ilya G.},
  title   = {Wndchrm -- An Open Source Utility for Biological Image Analysis},
  journal = {Source Code for Biology and Medicine},
  volume  = {3},
  pages   = {13},
  year    = {2008},
  doi     = {10.1186/1751-0473-3-13}
}

@article{lee2024neural,
  author  = {Lee, Jae Hee and Lanza, Sergio and Wermter, Stefan},
  title   = {From Neural Activations to Concepts: A Survey on Explaining Concepts in Neural Networks},
  journal = {Neurosymbolic Artificial Intelligence},
  year    = {2024},
  doi     = {10.3233/NAI-240743}
}

@article{widmer2023towards,
  author  = {Widmer, Cara Leigh and Sarker, Md Kamruzzaman and Nadella, Srikanth and Fiechter, Joshua and Juvina, Ion and Minnery, Brandon and Hitzler, Pascal and Schwartz, Joshua and Raymer, Michael},
  title   = {Towards Human-Compatible {XAI}: Explaining Data Differentials with Concept Induction over Background Knowledge},
  journal = {Journal of Web Semantics},
  volume  = {79},
  pages   = {100807},
  year    = {2023},
  doi     = {10.1016/j.websem.2023.100807}
}

@article{nanni2017handcrafted,
  title={Handcrafted vs. non-handcrafted features for computer vision classification},
  author={Nanni, Loris and Ghidoni, Stefano and Brahnam, Sheryl},
  journal={Pattern recognition},
  volume={71},
  pages={158--172},
  year={2017},
  publisher={Elsevier}
}

@inproceedings{
geirhos2018imagenettrained,
title={ImageNet-trained {CNN}s are biased towards texture; increasing shape bias improves accuracy and robustness.},
author={Robert Geirhos and Patricia Rubisch and Claudio Michaelis and Matthias Bethge and Felix A. Wichmann and Wieland Brendel},
booktitle={International Conference on Learning Representations},
year={2019},
url={https://openreview.net/forum?id=Bygh9j09KX},
}

@inproceedings{akkamahadevi2024automating,
  title={Automating CNN Neuron Interpretation using Concept Induction.},
  author={Akkamahadevi, Samatha Ereshi and Dalal, Abhilekha and Hitzler, Pascal},
  booktitle={ISWC (Posters/Demos/Industry)},
  year={2024}
}

\end{document}